\documentclass{article}

\PassOptionsToPackage{numbers, compress}{natbib}

     \usepackage[preprint]{neurips_2024}

\usepackage[utf8]{inputenc} 
\usepackage[T1]{fontenc}    
\usepackage{hyperref}       
\usepackage{url}            
\usepackage{booktabs}       
\usepackage{amsfonts}       
\usepackage{nicefrac}       
\usepackage{microtype}      
\usepackage{xcolor}         

\usepackage{graphicx}
\usepackage{array}
\usepackage{multirow}
\usepackage{subcaption}
\usepackage{algorithm}
\usepackage{algorithmic}
\usepackage{resizegather}
\usepackage{amsmath}
\usepackage{amssymb}
\usepackage{mathtools}
\usepackage{caption}
\usepackage{bbm}
\usepackage{nccmath}
\usepackage{amsfonts}
\usepackage{textcomp}

\graphicspath{ {./images/} }
\newcommand{\etal}{\emph{et al. }}

\title{Unbiased GNN Learning via\\ Fairness-Aware Subgraph Diffusion}

\author{
Abdullah Alchihabi, Yuhong Guo\\
School of Computer Science, Carleton University, Canada\\
\texttt{abdullahalchihabi@cmail.carleton.ca, yuhong.guo@carleton.ca}
}

\begin{document}

\maketitle

\begin{abstract}
Graph Neural Networks (GNNs) have demonstrated remarkable efficacy in tackling a wide array of graph-related tasks across diverse domains. However, a significant challenge lies in their propensity to generate biased predictions, particularly with respect to sensitive node attributes such as age and gender. 
These biases, inherent in many machine learning models, are amplified in GNNs due to the message-passing mechanism, which allows nodes to influence each other, rendering the task of making fair predictions notably challenging. This issue is particularly pertinent in critical domains 
where model fairness holds paramount importance. 
In this paper, we propose a novel generative Fairness-Aware Subgraph Diffusion (FASD) method 
for unbiased GNN learning. 
The method initiates by strategically sampling small subgraphs from the original large input graph, 
and then proceeds to conduct subgraph debiasing via generative fairness-aware graph diffusion processes
based on stochastic differential equations (SDEs).
To effectively diffuse unfairness in the input data, 
we introduce additional adversary bias perturbations to the subgraphs
	during the forward diffusion process, 
	and train score-based models to predict these applied perturbations, 
	enabling them to learn the underlying dynamics of the biases present in the data. 
Subsequently, the trained score-based models are utilized to further debias the original subgraph samples 
	through the reverse diffusion process. 
Finally, FASD induces fair node predictions on the input graph by performing standard GNN learning on the debiased subgraphs. 
Experimental results demonstrate the superior performance of the proposed method over state-of-the-art Fair GNN baselines across multiple benchmark datasets.
\end{abstract}

\section{Introduction}

Graph Neural Networks (GNNs) have 
exhibited remarkable efficacy in addressing a multitude of graph-related tasks across various domains, ranging from recommendation systems \cite{wu2019session} to drug discovery \cite{xiong2021graph}. Despite their successes,
GNNs have been show to be susceptible to inheriting biases present in the training data \cite{dai2021say}, potentially exhibiting more pronounced biases than traditional machine learning models. 
This increased vulnerability arises from GNNs' utilization of message passing mechanisms to propagate information across the graph, facilitating mutual influence among nodes and thereby 
amplifying existing biases in the data \cite{dai2021say}.
The prevalence of biased patterns in collected datasets \cite{beutel2017data,creager2019flexibly,dwork2012fairness}, coupled with standard GNNs' inability to effectively mitigate these biases \cite{dai2021say}, restricts their applicability in domains where fairness 
across sensitive attributes such as age, gender, and ethnicity 
is paramount, such as healthcare \cite{rajkomar2018ensuring} and job applicant evaluations \cite{mehrabi2021survey}.

Traditional fair learning methods \cite{kamiran2012data,zafar2017fairness} fail to consider the underlying graph structure and the interplay between nodes through message propagation, making them unsuitable for direct application to graph data. 
Therefore, it is important to develop methodologies capable of generating both accurate and fair predictions 
on graphs. 
Several endeavors have emerged to address this need, focusing on developing fairness-aware models tailored specifically to GNNs \cite{agarwal2021towards,dai2021say,kose2022fair} and balancing fairness and informativeness objectives in GNN training. 
In particular, adversarial
learning has been utilized to debias learned representations and predictions by employing adversary discriminator networks to remove sensitive attributes from learned embeddings \cite{dai2021say,bose2019compositional,fisher-etal-2020-debiasing}. 
Various strategies for fairness-aware graph data augmentation have been explored, including augmenting node features \cite{kose2022fair}, graph structures \cite{spinelli2021fairdrop}, or both \cite{ling2022learning} 
to debias the input data and promote fairness in downstream predictions with GNN models. 
Nevertheless, existing fairness-aware graph augmentation methods are often tailored to address specific forms of bias assumed to be present in the data \cite{kose2022fair,agarwal2021towards,spinelli2021fairdrop},
lacking sufficient and broad applicability
across the diverse application domains of GNNs \cite{ling2022learning}, 
highlighting the necessity for developing data-adaptive fairness-aware graph augmentation and learning methods.

In this paper, we introduce a novel generative method, named as Fairness-Aware Subgraph Diffusion (FASD),
to tackle the challenging task of fair GNN learning for node classification. 
FASD first strategically samples small subgraphs from the original large input graph, 
then conducts subgraph debiasing via generative fairness-aware graph diffusion processes
based on stochastic differential equations (SDEs),
and finally induces fair node predictions on the input graph by performing standard GNN learning on the debiased subgraphs. 
In particular, to learn the SDE-based graph diffusion model for mitigating existing bias or unfairness in the input data, 
we first perturb the subgraphs---both the node features and the connectivity structures---in the forward diffusion process 
with additional adversary bias based on a learned sensitive attribute prediction model. 
Subsequently, 
we employ score-based models to estimate these applied perturbations, 
enabling them to adaptively learn the intrinsic bias patterns present in the data. 
The trained score-based models are then leveraged to debias the original sampled subgraphs 
through a reverse diffusion process to obtain debiased subgraph samples. 
We conduct experiments on several benchmark graph datasets for fair GNN learning. 
Experimental results validate the efficacy of our proposed method, showcasing its superior performance compared to state-of-the-art fair GNN baselines across the benchmarks. 
The main contributions of this work are twofold: 
(1) The proposed Fairness-Aware Subgraph Diffusion method represents 
a pioneering effort in fairness-aware graph diffusion, utilizing the diffusion process to debias subgraph instances and promote fairness in downstream tasks. 
(2) Our empirical findings on diverse benchmark datasets underscore the efficacy of our proposed method in eliminating biases inherent in the input data, 
leading to state-of-the-art fairness outcomes for GNN learning.

\section{Related Works}
\subsection{Fair GNN Learning}
Existing fair GNN learning methods can be classified into three primary categories: pre-processing, in-processing, and post-processing techniques. 
Pre-processing methods aim to mitigate bias within the input graph data prior to training standard GNN models for downstream tasks. 
For instance, FairWalk \cite{rahman2019fairwalk} modifies the random walk component in node2vec \cite{grover2016node2vec} to sample more diverse neighborhood information, resulting in less biased node embeddings. FairDrop \cite{spinelli2021fairdrop} employs a heuristic edge-dropout strategy to debias the graph structure by removing edges between nodes with similar sensitive attribute values. Graphair \cite{ling2022learning} filters out sensitive attribute information from graph structure and node features to learn fair graph representations. 
EDITS \cite{dong2022edits} introduces a Wasserstein distance approximator that alternates between debiasing node features and graph structure. 
However, many existing pre-processing methods lack sufficient robustness, 
as they are specifically designed to address particular forms of bias presumed to exist in the data, 
which restricts their applicability across diverse domains. 
In-processing methods adapt the learning process to achieve a balance between accuracy and fairness through the incorporation of fairness-aware regularization techniques. 
In particular, 
NIFTY \cite{agarwal2021towards} introduces graph contrastive learning with fairness-aware augmentations and a novel loss function to balance fairness and stability. FairAug \cite{kose2022fair} proposes several fairness-aware adaptive graph augmentation strategies, including feature masking, node sampling, and edge deletion, to debias the graph data. REDRESS \cite{dong2021individual} introduces a ranking-based objective to promote individual fairness. Adversarial learning has also been utilized to eliminate sensitive attribute information from the learned node embeddings \cite{fisher-etal-2020-debiasing,dai2021say,bose2019compositional}. However, in-processing methods often require careful training to balance fairness and informativeness objectives, which can significantly compromise their stability. 
Post-processing methods on the other hand focus on modifying the predictions of GNN models to promote fairness and alleviate bias \cite{hardt2016equality,pleiss2017fairness}. Further details on Fair GNN Learning methods can be found in the survey paper \cite{choudhary2022survey}.

\subsection{Graph Diffusion Methods}

Graph diffusion methods have exhibited notable efficacy in capturing the underlying distribution of graph data, facilitating the synthesis of high-fidelity graph samples closely aligned with the input data distribution. These methods are broadly categorized into two main groups: continuous diffusion methods and discrete diffusion methods. 
Discrete methods often model the diffusion process using Markov noise processes, while stochastic differential equations commonly model the diffusion process in continuous methods. 
For instance, DiGress \cite{vignac2022digress} employs a discrete diffusion process to generate graph instances with categorical node features and edge attributes. GDSS \cite{jo2022score}, on the other hand, models the joint distribution of node features and graph structure using stochastic differential equations. Kong \etal \cite{kong2023autoregressive} accelerate the training and sampling processes of diffusion models through an autoregressive diffusion model, while Haefeli \etal \cite{haefeli2022diffusion} leverage discrete noise in the forward Markov process to ensure the discreteness of the graph structure throughout the forward diffusion process, thereby enhancing sample quality and speeding up sampling. Additionally, PRODIGY \cite{sharma2023plug} adopts projected diffusion to enforce explicit constraints on synthesized graph samples in terms of edge count and valency. Further details on graph diffusion methods can be found in \cite{liu2023generative}. 
However, prior graph diffusion methods typically synthesize new samples from the distribution of input graph data, rendering them susceptible to inheriting biases present in the input data. To our knowledge, our Fairness-Aware Subgraph Diffusion method represents the first endeavor towards a fairness-aware graph diffusion approach, harnessing graph diffusion to promote fairness in downstream tasks.

\section{Method}
\subsection{Problem Setup}

We consider the following fairness-sensitive semi-supervised node classification problem. 
The input graph is given as $G=(V,E)$, where $V$ is the set of nodes with size $N=|V|$ 
and $E$ denotes the set of edges of the graph. Typically, $E$ is represented by an adjacency matrix $A$ of size $N \times N$. Each node in the graph is associated with a feature vector of size $D$, 
and the features of all nodes in the graph are represented by an input feature matrix $X\in\mathbb{R}^{N\times D}$. Additionally, each node in the graph is associated with a binary sensitive attribute $s\in \{0,1\}$, and the sensitive attributes of all nodes are represented 
by a sensitive attribute vector $S$ of size $N$. The nodes 
are split into two distinct subsets: $V_{\ell}$ and $V_u$ comprising the labeled and unlabeled nodes, respectively. The labels of nodes in $V_{\ell}$ are represented by a label indicator matrix $Y^{\ell} \in \{0,1\}^{N_{\ell} \times C}$, where $C$ is the number of classes and $N_{\ell}$ is the number of labeled nodes in the graph. 
The objective is to learn a GNN model capable of accurately and fairly predicting the labels of the nodes. 
Here, we focus on a widely studied fairness criterion, group fairness \cite{hardt2016equality,zemel2013learning},
which stipulates that a model should not exhibit bias in favor of or against demographic groups in the data, as defined by their sensitive attributes. 
To assess the fairness of a model, we employ two standard metrics: statistical parity \cite{dwork2012fairness} and equal opportunity \cite{hardt2016equality}. 

\subsection{Fairness-Aware Subgraph Diffusion}

\begin{figure}[t]
    \centering 
    \includegraphics[width=\textwidth]{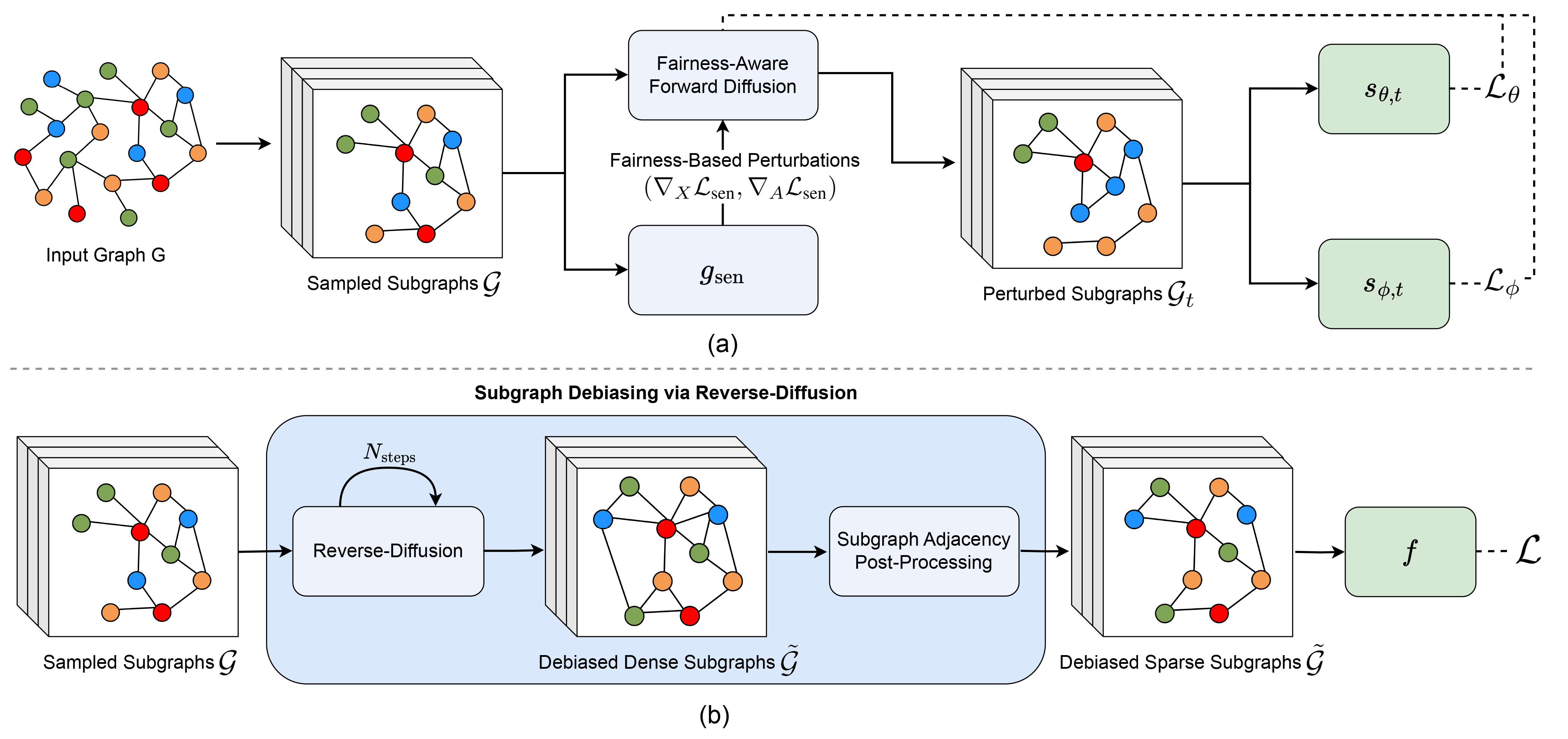}
    \caption{Overview of the proposed FASD method. \textbf{(a) Fairness-Aware Forward Diffusion Process.} Subgraphs $\mathcal{G}$ sampled from the input graph $G$ undergo stochastic and fairness-based perturbations within the forward
	subgraph diffusion process. Score-based models $s_{\theta,t}$ and $s_{\phi,t}$ are trained to approximate these perturbations by minimizing losses $\mathcal{L}_{\theta}$ and $\mathcal{L}_{\phi}$, respectively. \textbf{(b) Subgraph Debiasing and fair node classification.} The sampled subgraphs $\mathcal{G}$ are debiased via reverse diffusion to obtain dense debiased subgraphs $\tilde{\mathcal{G}}$, which are then sparsified during post-processing. Finally, node classification model $f$ is trained on $\tilde{\mathcal{G}}$ to minimize node classification loss $\mathcal{L}$.}
    \label{fig:diagram}
\end{figure}

In this section, we present the proposed 
Fairness-Aware Subgraph Diffusion (FASD) method designed to address the intricate challenge of fair semi-supervised node classification. 
The generative continuous diffusion processes for FASD are 
modeled on a set of subgraph instances sampled from the input graph
via Stochastic Differential Equations (SDEs). 
During the forward diffusion process, 
fairness-based adversary perturbations, derived from a sensitive attribute prediction model, 
are introduced onto the subgraphs.
These perturbations serve to amplify existing biases within subgraph instances and furnish learnable debiasing targets for score-based bias prediction models. The trained score-based models are then employed to mitigate biases on the input subgraph instances via the reverse-diffusion process, resulting in their debiased counterparts. Finally, standard graph neural network (GNN) learning is applied to the debiased subgraphs, 
facilitating the generation of node predictions that are both fair and accurate.
An overview of the FASD framework is depicted in Figure \ref{fig:diagram}. 
In the remainder of this section, we elaborate on the details of the proposed FASD method.

\subsubsection{Subgraph-level Instance Sampling}
In the context of fair semi-supervised node classification, the input comprises a single graph $G$. 
However, generative graph diffusion methods require a multitude of graph samples to effectively capture the underlying data distribution. 
Furthermore, graph diffusion methods typically operate on relatively small graphs 
due to their computational cost. Conversely, the input graph $G$ in fair semi-supervised node classification can be very large, consisting of tens of thousands of nodes. 
Hence, to reconcile this disparity 
and ensure the scalability of our proposed FASD method to large graph datasets, 
we represent the input graph $G$ as a set of small subgraph instances $\mathcal{G}=\{G^{(i)}\}_{i=1}^{M}$.

In order to generate the set 
of subgraphs $\mathcal{G}$ that can effectively represent the input graph $G$,
we adopt a localized subgraph sampling strategy 
denoted as follows:
\begin{equation}
    G^{(i)} = \text{Subgraph\_Sampling}(G, u, d, k)
\end{equation}
where
$u$ denotes the starting node for the sampling process, $d$ signifies the depth of the subgraph measured in hops, and $k$ represents the number of neighbors to sample for each node at each hop in $G^{(i)}$. Specifically, $G^{(i)}$ is constructed by initiating the sampling process at node $u$ and iteratively sampling $k$ neighboring nodes at each hop, expanding $G^{(i)}$ in a breadth-first manner until the subgraph reaches a depth of $d$. Further details on the sampling procedure are provided in the Appendix \ref{section:subgraph_sampling}.

\subsubsection{Fairness-Aware Forward Diffusion}

The forward graph diffusion process is modeled via a system of Stochastic Differential Equations (SDEs) \cite{jo2022score,song2020score}, as follows: 
\begin{equation}
\label{eq:forward_diff}
    \text{d}G_{t}^{(i)} = \text{f}_t(G_{t}^{(i)}) \text{d}t + \text{g}_t(G_{t}^{(i)}) \text{d}\mathbf{w}, \quad G_{0}^{(i)} \in \mathcal{G}.
\end{equation}
Here, $G_{0}^{(i)} \in \mathcal{G}$ denotes the input subgraph, and $G_{t}^{(i)}$ represents the perturbed subgraph at time-step $t$, 
where $t \in [0,T]$. $\text{f}_t(.)$ denotes the linear drift coefficient, $\text{g}_t$ represents the diffusion coefficient, and 
$\mathbf{w}$ signifies the Wiener process. 
Previous graph diffusion methods 
are fairness-agnostic, 
perturbing input samples solely with stochastic noise during the forward diffusion process. 
With the goal of inducing a debiasing reverse diffusion process for fair graph generation, 
we propose to further incorporate fairness-adversary perturbations in the forward diffusion process. 

As the key aspect of fairness is reflected in whether a prediction model exhibits bias in favor of or
against demographic groups defined by the sensitive attributes of the data, 
we attempt to capture the intrinsic bias or unfairness patterns present in the input graph data 
with a sensitive attribute prediction model. 
Specifically, we introduce a GNN-based sensitive attribute prediction model, $g_{\text{sen}}$, 
to predict the sensitive attribute values of the nodes in each generated subgraph $G^{(i)}$ 
based on the node features $X^{(i)}$ and adjacency matrix $A^{(i)}$ as follows:
\begin{equation}
    \hat{S}^{(i)} = g_{\text{sen}}(X^{(i)}, A^{(i)}).
    \label{eq:g_sen}
\end{equation}
The sensitive attribute predictor $g_{\text{sen}}$ is trained on all the subgraphs in $\mathcal{G}$ to minimize the following cross-entropy loss: 
\begin{equation}
	\mathcal{L}^{\text{total}}_{\text{sen}} =  \sum\nolimits_{G^{(i)} \in \mathcal{G}} \mathcal{L}_{\text{sen}}(X^{(i)},A^{(i)}) = \sum\nolimits_{G^{(i)} \in \mathcal{G}} \sum\nolimits_{u \in V^{(i)}} \ell_{ce}(\hat{s}^{(i)}_u,s^{(i)}_u) 
   \label{eq:sen_att_loss}
\end{equation}
where $\hat{s}^{(i)}_u$ denotes the predicted sensitive attribute value for node $u$ and 
$s^{(i)}_u$ is the corresponding ground-truth sensitive attribute value; 
$\ell_{ce}(\cdot,\cdot)$ is the standard cross-entropy loss.  
A lower sensitivity attribute prediction loss $\mathcal{L}_{\text{sen}}$
indicates higher bias, meaning lower fairness, within the given subgraph, and vice versa.
The trained sensitive attribute prediction model $g_{\text{sen}}$ is then 
leveraged to generate fairness-based adversary perturbations by utilizing the gradients of its loss $\mathcal{L}_{\text{sen}}$. 
Note the bias within $G^{(i)}$, reflected in $X^{(i)}$ and $A^{(i)}$, 
can be amplified or diminished by subtracting or adding the gradients $\nabla_{X}\mathcal{L}_{\text{sen}}$ and $\nabla_{A}\mathcal{L}_{\text{sen}}$ 
to perturb $X^{(i)}$ and $A^{(i)}$, respectively.

In particular, 
within the proposed fairness-aware forward diffusion process,  
we perturb the node features $X^{(i)}$ and adjacency matrix $A^{(i)}$ of each subgraph $G_{0}^{(i)}\in\mathcal{G}$ 
with fairness-based and stochastic perturbations as follows:
\begin{equation}
\begin{aligned}
    &   {X}^{(i)}_t = \mu_{t}(X^{(i)}_0) + \sigma_{t}(X^{(i)}_0) \times \epsilon_X - \gamma_X \nabla_{X}\mathcal{L}_{\text{sen}}(X^{(i)}_0,A^{(i)}_0)    \\
    &  {A}^{(i)}_t = \mu_{t}(A^{(i)}_0)  + \sigma_{t}(A^{(i)}_0)\times \epsilon_A   - \gamma_A \nabla_{A}\mathcal{L}_{\text{sen}}(X^{(i)}_0,A^{(i)}_0). 
\end{aligned}
\label{eq:forward_sub_pert}
\end{equation}
Here 
$\epsilon_X \sim \mathcal{N}(0, 1)^{(N^{(i)}, D)}$ and $\epsilon_A \sim \mathcal{N}(0, 1)^{(N^{(i)}, N^{(i)})}$ denote stochastic Gaussian perturbations. 
$\mu_{t}(X_0)$ and $\sigma_{t}(X_0)$ represent the mean and standard deviation of $X_0$ at time $t$, respectively. 
Similarly, $\mu_{t}(A_0)$, $\sigma_{t}(A_0)$ denote the mean and standard deviation of $A_0$ at time $t$.
These means and standard deviations are calculated using standard perturbation kernels (details are provided in Appendix \ref{section:kernels}), 
which govern the evolution of ${X}^{(i)}$ and ${A}^{(i)}$ over time during the perturbation process. 
As time $t$ progresses, the means are scaled down and standard deviations are amplified, reflecting the gradual perturbation of ${X}^{(i)}$ and ${A}^{(i)}$. 
The terms $\nabla_{X}\mathcal{L}_{\text{sen}}$ and $\nabla_{A}\mathcal{L}_{\text{sen}}$ represent the fairness-based perturbations generated using model $g_{\text{sen}}$. 
The coefficients $\gamma_X$ and $\gamma_A$ control the relative contributions of the fairness-based perturbations to the stochastic perturbations,
we determine their values as follows:
\begin{equation}
    \gamma_X = \lambda_X \frac{\|\sigma_{t}(X^{(i)}_0) \times \epsilon_X\|_2^2  }
	{ \| \nabla_{X}\mathcal{L}_{\text{sen}}(X^{(i)}_0,A^{(i)}_0)\|_2^2  }, \quad 
	\gamma_A = \lambda_A \frac{\|\sigma_{t}(A^{(i)}_0)\times \epsilon_A\|_2^2  }{ \| \nabla_{A}\mathcal{L}_{\text{sen}}(X^{(i)}_0,A^{(i)}_0)\|_2^2  }
    \label{eq:gamma}
\end{equation}
where $\lambda_X$ and $\lambda_A$ are scalar hyper-parameters controlling the scale of the fairness-aware perturbations applied to the subgraph node features ${X}^{(i)}_t$ and subgraph adjacency matrix ${A}^{(i)}_t$, respectively. By scaling $\gamma_X$ and $\gamma_A$ based on the ratio of the norm of the stochastic perturbations to the norm of the fairness-based perturbations, we ensure a 
stable and well balanced forward diffusion process.

\subsubsection{Estimating Diffusion Perturbations via Score-Based Models}
To estimate the perturbations in the fairness-aware forward diffusion process, 
we introduce two score-based models: $s_{\theta,t}(G^{(i)}_t)$ and $s_{\phi,t}(G^{(i)}_t)$. 
These models are designed to approximate the perturbation scores applied to obtain $X^{(i)}_t$ and $A^{(i)}_t$, respectively. 
These models leverage permutation-equivariant GNN-based architectures \cite{jo2022score}, given their effectiveness at capturing the complex 
interdependencies between $X^{(i)}_t$ and $A^{(i)}_t$ over time. 
Specifically, the score-based model $s_{\theta,t}(G^{(i)}_t)$ is defined as follows:
\begin{equation}
\begin{aligned}
    &   s_{\theta,t}(G^{(i)}_t) = \text{MLP}_{X}([\{H_j\}_{j=0}^{L}])    \\
    &   H_{j+1}=\text{GNN}_{X}(H_{j},A^{(i)}_t), \quad H_{0}=X^{(i)}_t 
\end{aligned}
\label{eq:s_theta}
\end{equation}
where $L$ denotes the number of GNN-based layers, $\text{MLP}$ signifies a multi-layer perceptron (fully-connected layers), and $[.]$ represents the concatenation operation. 
The output of $s_{\theta,t}(G^{(i)}_t) \in \mathbb{R}^{(N^{(i)},D)}$ has the same dimensionality as $X^{(i)}_t$. 
Similarly, the score-based model $s_{\phi,t}(G^{(i)}_t)$
estimates the perturbation score applied on the graph structure, and is defined as: 
\begin{equation}
\begin{aligned}
    &   s_{\phi,t}(G^{(i)}_t) = \text{MLP}_{A}([\{\text{GMH}(H_j,(A^{(i)}_t)^{p} )\}_{j=0,p=1}^{K,P}])    \\
    &   H_{j+1}=\text{GNN}_{A}(H_{j},A^{(i)}_t), \quad H_{0}=X^{(i)}_t
\end{aligned}
\label{eq:s_phi}
\end{equation}
where $\text{GMH}$ denotes the Graph Multi-Headed attention (GMH) module \cite{baek2021accurate}, $(A^{(i)}_t)^{p}$ represents the higher-order subgraph adjacency matrix. Here, $K$ and $P$ denote the number of GMH layers and the highest-order of adjacency matrix, respectively. By leveraging higher-order subgraph adjacency matrices, $s_{\phi,t}(G^{(i)}_t)$ can capture long-range interactions among nodes within the subgraph. 

To introduce time dependence, the outputs of $s_{\theta,t}(G^{(i)}_t)$ and $s_{\phi,t}(G^{(i)}_t)$ are scaled by $\sigma_{t}(X^{(i)}_0)$ and $\sigma_{t}(A^{(i)}_0)$, respectively \cite{song2020score,jo2022score}. Consequently, $s_{\theta,t}(G^{(i)}_t)$ and $s_{\phi,t}(G^{(i)}_t)$ are trained to estimate the $\sigma_{t}$-scaled perturbations applied to obtain $X^{(i)}_t$ and $A^{(i)}_t$ by minimizing the following loss functions:
\begin{equation}
\begin{aligned}
    & \mathcal{L}_{\theta} = \mathbb{E}_t \{ \mathbb{E}_{G^{(i)}_0} \mathbb{E}_{G^{(i)}_t|G^{(i)}_0}  
	\| s_{\theta,t}(G^{(i)}_t) - \epsilon_X + \frac{\gamma_X}{\sigma_t(X^{(i)}_0)}   \nabla_{X}\mathcal{L}_{\text{sen}}(X^{(i)}_0,A^{(i)}_0)    \|_2^2    \}   \\
    & \mathcal{L}_{{\phi}} = \mathbb{E}_t \{ \mathbb{E}_{G^{(i)}_0} \mathbb{E}_{G^{(i)}_t|G^{(i)}_0}  
	\| s_{\phi,t}(G^{(i)}_t) -  \epsilon_A   + \frac{\gamma_A}{\sigma_t(A^{(i)}_0)}   \nabla_{A}\mathcal{L}_{\text{sen}}(X^{(i)}_0,A^{(i)}_0) \|_2^2    \}
\end{aligned}
\end{equation}
Here $G^{(i)}_0$ is sampled from $\mathcal{G}$. The expectations are computed using Monte Carlo estimates by sampling $(t,G^{(i)}_0,G^{(i)}_t)$. 
Through this training process, the score-based models learn to approximate the perturbations applied in the fairness-aware forward diffusion process, thus enabling them to capture the intrinsic bias patterns present in the data. 
The details of the training procedure of the score-based models are provided in Appendix \ref{section:score_training}.

\subsubsection{Subgraph Debiasing via Reverse Diffusion}

Prior graph diffusion methods synthesize new graph samples from random noise 
through reverse diffusion
via reverse-time SDEs \cite{anderson1982reverse}. 
In contrast, we utilize 
the score-based models,   
$s_{\theta,t}$ and $s_{\phi,t}$, learned
from the forward diffusion process to estimate perturbations and further reduce the bias patterns 
in the input subgraph instances through a fairness-aware reverse diffusion process,  
thereby generating their debiased counterparts. 
This reverse graph diffusion process over a subgraph $G_t^{(i)}=(X^{(i)}_t,A^{(i)}_t)$ 
can be modeled 
via the following reverse-time SDEs:
\begin{equation}
\begin{aligned}
    & \text{d}X^{(i)}_t = [\text{f}_{1,t}(X^{(i)}_t) - \text{g}_{1,t}^2 s_{\theta,t}(G^{(i)}_t) ] \text{d}\Bar{t} + \text{g}_{1,t} \text{d}\bar{\mathbf{w}}_{1} \\
    & \text{d}A^{(i)}_t = [\text{f}_{2,t}(A^{(i)}_t) - \text{g}_{2,t}^2 s_{\phi,t}(G^{(i)}_t) ] \text{d}\Bar{t} + \text{g}_{2,t} \text{d}\bar{\mathbf{w}}_{2}
\end{aligned}
\end{equation}
Here $\Bar{\mathbf{w}}_{1}$ and $\Bar{\mathbf{w}}_{2}$ 
represent the reverse-time Wiener processes, while $\text{d}\bar{t}$ denotes the infinitesimal negative time step. 
$\text{f}_{1,t}$ and $\text{f}_{2,t}$ are linear drift coefficients, and $\text{g}_{1,t}$ and $\text{g}_{2,t}$ are diffusion coefficients. 

Specifically, 
instead of starting from random noise, 
we initialize the reverse diffusion process with the subgraphs $\mathcal{G}$ sampled
from the input data distribution---the original input graph $G$. 
The score-based models are then deployed to iteratively debias the subgraph node features and connectivity structure at each step of the reverse diffusion process, 
thereby approximating an unbiased distribution of the input data.
This reverse diffusion process can be denoted as follows:
\begin{equation}
    \tilde{G}^{(i)} = \text{reverse\_diffusion}({G}^{(i)}, N_{\text{steps}})
\end{equation}
where $\tilde{G}^{(i)}$ represents the debiased subgraph for the input subgraph ${G}^{(i)}$, and $N_{\text{steps}}$ denotes the number of steps of the reverse-diffusion process. 
As the reverse-diffusion process is modeled as a system of two SDEs interconnected through score functions, we utilize the Predictor-Corrector (PC) Sampler \cite{song2020score} to approximate this process. The PC sampler comprises four components: two Predictors ($\text{Predictor}_{X}$, $\text{Predictor}_{A}$) and two Correctors ($\text{Corrector}_{X}$, $\text{Corrector}_{A}$). 
These components iteratively debias the input subgraph instances. At each reverse-diffusion step, the Predictors employ current estimates of subgraph node features and adjacency matrix, leveraging the trained score-based models $s_{\theta,t}$ and $s_{\phi,t}$ to generate initial debiased estimates. Subsequently, the Correctors refine these estimates. 
After $N_{\text{steps}}$ reverse-diffusion steps, we obtain the debiased subgraph node input features $\tilde{X}^{(i)}$ and the corresponding debiased adjacency matrix $\tilde{A}^{(i)}$.  
Additional details on the adopted PC sampler are provided in Appendix \ref{section:sampler}.

After reverse diffusion, we further post-process the debiased subgraph adjacency matrix $\tilde{A}^{(i)}$ 
to reduce noise by pruning its weak connections as follows: 
\begin{equation}
    \tilde{A}^{(i)}[u,v]= \begin{cases}
    \tilde{A}^{(i)}[u,v],& \text{if } \, \tilde{A}^{(i)}[u,v]\geq \tau \\
    0,              & \text{otherwise}.
\end{cases}
\end{equation}
Here $\tau$ denotes the threshold for pruning weak edges. 
This post-processing step effectively filters out noise from the edge weights and sparsifies the debiased subgraph connectivity. 
By applying the reverse-diffusion process and subsequent post-processing mechanism to each subgraph ${G}^{(i)} \in \mathcal{G}$, we obtain a corresponding set of debiased subgraph instances $\tilde{\mathcal{G}}=\{ \tilde{G}^{(i)} \}_{i=1}^{M}$.

\subsubsection{Fair Node Classification with Debiased Subgraphs}
The debiased subgraph set $\mathcal{\tilde{G}}$, obtained through our proposed FASD, 
represents the debiased input graph and 
serves as the training data for a standard GNN-based node classification model $f$, 
such that
\begin{equation}
    P^{(i)} = f(\tilde{X}^{(i)},\tilde{A}^{(i)}).
\end{equation}
Here $\tilde{X}^{(i)}$ and $\tilde{A}^{(i)}$ denote the debiased node feature matrix and debiased adjacency matrix of subgraph $\tilde{G}^{(i)} \in \mathcal{\tilde{G}}$, respectively. 
$P^{(i)} \in \mathbb{R}^{N^{(i)}\times C}$ represents the predicted class probability matrix for all nodes in the subgraph. 
The GNN model $f$ is trained by minimizing the node classification loss across all the debiased subgraphs:
\begin{equation}
	\mathcal{L} = \sum\nolimits_{\tilde{G}^{(i)} \in \mathcal{\tilde{G}}} \, \sum\nolimits_{u \in V_{\ell}^{(i)}} \ell_{\text{ce}}(P^{(i)}_u, Y^{\ell}_u)
\end{equation}
where
$\ell_{\text{ce}}$ denotes the standard cross-entropy loss, while $P^{(i)}_u$ and $Y^{\ell}_u$
are the predicted class probability vector and ground-truth label indicator vector of node $u$, respectively. $V_{\ell}^{(i)}$ denotes the set of labeled nodes in subgraph $\tilde{G}^{(i)}$. 
For inference on unlabeled nodes, 
the predicted class probability vector $P_u$ for a given node $u$ is obtained 
by averaging the prediction vectors of model $f$ 
over all the subgraphs in $\mathcal{\tilde{G}}$ containing node $u$: 
\begin{equation}
    P_u = \frac{1}{|\mathcal{\tilde{G}}_u|}   \sum\nolimits_{\tilde{G}^{(i)} \in \mathcal{\tilde{G}}_u} \,  P^{(i)}_u 
\end{equation}
where $P^{(i)}_u$ is predicted class probability vector of node $u$ using subgraph $\tilde{G}^{(i)}$, and 
$\mathcal{\tilde{G}}_u$ signifies the set of subgraphs from $\mathcal{\tilde{G}}$ containing node $u$. 

\section{Experiments}

\subsection{Experimental Setup}
\label{section:Experimental_Setup}

\paragraph{Datasets \& Baselines}
We conduct experiments on three benchmark graph datasets: NBA \cite{dai2021say}, Pokec-z and Pokec-n \cite{takac2012data}. 
Across all three datasets, the sensitive attributes and class labels take binary values. 
To ensure a fair comparison, we adhere to the same train/validation/test splits provided by \cite{ling2022learning}. 
Comprehensive details regarding the benchmark datasets are provided in Appendix \ref{section:dataset}. 
We compare our proposed method with
five fair GNN learning methods, 
namely FairWalk \cite{rahman2019fairwalk}, FairDrop \cite{spinelli2021fairdrop}, NIFTY \cite{agarwal2021towards}, FairAug \cite{kose2022fair} and Graphair \cite{ling2022learning}, along with two fairness-agnostic graph contrastive learning methods, GRACE \cite{zhu2020deep}, and GCA \cite{zhu2021graph}.

\paragraph{Evaluation Criteria}
We evaluate the performance of our proposed FASD and all comparison baselines using three distinct metrics: accuracy, demographic parity, and equal opportunity.
Demographic parity is computed as follows: $\Delta_{DP} = | P(\hat{Y}=1|S=0) - P(\hat{Y}=1|S=1)|$ where $\hat{Y}$ denotes the predicted class label 
and $S$ denotes the sensitive attribute value. 
Meanwhile, equal opportunity is calculated as follows: $\Delta_{EO} = | P(\hat{Y}=1|S=0,Y=1) - P(\hat{Y}=1|S=1,Y=1)|$ where $Y$ denotes the ground-truth class label. 
All three metrics are measured over the test nodes of each dataset. It is important to note 
that smaller values for $\Delta_{DP}$ and $\Delta_{EO}$ 
indicate enhanced fairness.

\begin{table}[t]
\caption{The overall performance (standard deviation within brackets) on NBA, Pokec-z and Pokec-n datasets. The best results are indicated in bold font and the second best results are underlined.}
\setlength{\tabcolsep}{1pt}
\resizebox{\textwidth}{!}{ 
\begin{tabular}{l|ccc||ccc||ccc}
\hline
& \multicolumn{3}{c||}{NBA}   & \multicolumn{3}{c||}{Pokec-z}                            & \multicolumn{3}{c}{Pokec-n}                            \\
& \multicolumn{1}{l}{Acc.\%}   & \multicolumn{1}{l}{$\Delta_{\text{DP}}$\%}    & \multicolumn{1}{l||}{$\Delta_{\text{EO}}$\%}  
& \multicolumn{1}{l}{Acc.\%}   & \multicolumn{1}{l}{$\Delta_{\text{DP}}$\%}    & \multicolumn{1}{l||}{$\Delta_{\text{EO}}$\%}   
& \multicolumn{1}{l}{Acc.\%}   & \multicolumn{1}{l}{$\Delta_{\text{DP}}$\%}    & \multicolumn{1}{l}{$\Delta_{\text{EO}}$\%}   \\
\hline
\hline                          
FairWalk      & ${64.54}_{(2.35)}$      & ${3.67}_{(1.28)}$ & ${9.12}_{(7.06)}$  & ${67.07}_{(0.24)}$ & ${7.12}_{(0.74)}$ & ${8.24}_{(0.75)}$  & ${65.23}_{(0.78)}$   & ${4.45}_{(1.25)}$ & ${4.59}_{(0.86)}$                       \\
FairWalk+X    & ${69.74}_{(1.71)}$ & ${14.61}_{(4.98)}$ & ${12.01}_{(5.38)}$  & ${69.01}_{(0.38)}$ & ${7.59}_{(0.96)}$ & ${9.69}_{(0.09)}$    & ${67.65}_{(0.60)}$   & ${4.46}_{(0.38)}$ & ${6.11}_{(0.54)}$                \\
GRACE         & $\underline{70.14}_{(1.40)}$ & ${7.49}_{(3.78)}$ & ${7.67}_{(3.78)}$  & ${68.25}_{(0.99)}$ & ${6.41}_{(0.71)}$ & ${7.38}_{(0.84)}$   & $\underline{67.81}_{(0.41)}$   & ${10.77}_{(0.68)}$  & ${10.69}_{(0.69)}$                  \\
GCA           & $\mathbf{70.43}_{(1.19)}$ & ${18.08}_{(4.80)}$ & ${20.04}_{(4.34)}$  & $\mathbf{69.34}_{(0.20)}$ & ${6.07}_{(0.96)}$ & ${7.39}_{(0.82)}$   & ${67.07}_{(0.14)}$   & ${7.90}_{(1.10)}$  & ${8.05}_{(1.07)}$                \\
FairDrop      & ${69.01}_{(1.11)}$ & ${3.66}_{(2.32)}$ & ${7.61}_{(2.21)}$  & ${67.78}_{(0.60)}$ & ${5.77}_{(1.83)}$ & ${5.48}_{(1.32)}$   & ${67.32}_{(0.61)}$   & ${4.05}_{(1.05)}$  & ${3.77}_{(1.00)}$                \\
NIFTY         & ${69.93}_{(0.09)}$ & ${3.31}_{(1.52)}$ & ${4.70}_{(1.04)}$  & ${67.15}_{(0.43)}$ & ${4.40}_{(0.99)}$ & ${3.75}_{(1.04)}$  & ${65.52}_{(0.31)}$   & ${6.51}_{(0.51)}$  & ${5.14}_{(0.68)}$                 \\
FairAug       & ${66.38}_{(0.85)}$ & ${4.99}_{(1.02)}$ & ${6.21}_{(1.95)}$  & $\underline{69.17}_{(0.18)}$ & ${5.28}_{(0.49)}$ & ${6.77}_{(0.45)}$ & $\mathbf{68.61}_{(0.19)}$   & ${5.10}_{(0.69)}$  & ${5.22}_{(0.84)}$                  \\
Graphair      & ${69.36}_{(0.45)}$ & $\underline{2.56}_{(0.41)}$ & $\underline{4.64}_{(0.17)}$  & ${68.17}_{(0.08)}$ & $\mathbf{2.10}_{(0.17)}$ & $\underline{2.76}_{(0.19)}$  & ${67.43}_{(0.25)}$   & $\underline{2.02}_{(0.40)}$  & $\underline{1.62}_{(0.47)}$         \\
FASD  & ${69.22}_{(0.96)}$ & $\mathbf{0.92}_{(0.66)}$   & $\mathbf{4.47}_{(0.23)}$  & ${66.15}_{(0.49)}$ & $\underline{2.28}_{(0.42)}$  & $\mathbf{1.96}_{(0.51)}$  &  ${66.34}_{(0.37)}$  & $\mathbf{0.79}_{(0.38)}$   & $\mathbf{0.91}_{(0.53)}$      \\

\hline
\end{tabular} }
\label{table:comparison_results}
\vskip -0.20in
\end{table}

\subsection{Comparison Results}
\label{section:Comparison_Results}

We evaluate the efficacy of our proposed FASD method 
and compare it with the other comparison methods
in the context of fair semi-supervised node classification. The mean and standard deviation of all three performance metrics—Accuracy (Acc.\%), Demographic Parity ($\Delta_{\text{DP}}$\%), and Equal Opportunity ($\Delta_{\text{EO}}$\%)—are reported over 5 runs in Table \ref{table:comparison_results}.

The table clearly shows the trade-off between accuracy and fairness metrics, where methods achieving the highest accuracy tend to exhibit pronounced bias (low fairness). 
Conversely, approaches prioritizing fairness often experience a decline in accuracy. This underscores the formidable challenge in fair node classification, necessitating the enhancement of fairness without significantly compromising accuracy.
Our proposed method achieves the most notable results in terms of equal opportunity across all three datasets, significantly surpassing all other methods. Notably, it yields performance gains of 29\% (0.8) and 43\% (0.71) over the second-best method on the Pokec-z and Pokec-n datasets, respectively. 
Similarly, our FASD method outperforms all baselines in demographic parity on NBA and Pokec-n datasets, while securing the second-best results on Pokec-z dataset. The substantial improvement in demographic parity over the second-best baseline exceeds 64\% (1.64) and 60\% (1.23) on NBA and Pokec-n datasets, respectively.
Furthermore, FASD demonstrates only a minimal decrease in accuracy, underscoring its ability to strike 
a great balance between fairness and accuracy. These findings highlight the superior performance of our proposed method compared to existing state-of-the-art fair GNN learning methods.

\begin{table}[t]
\caption{Ablation study results (standard deviation within brackets) on NBA, Pokec-z and Pokec-n.}
\setlength{\tabcolsep}{1pt}
\resizebox{\textwidth}{!}{ 
\begin{tabular}{l|ccc||ccc||ccc}
\hline
& \multicolumn{3}{c||}{NBA}   & \multicolumn{3}{c||}{Pokec-z}                            & \multicolumn{3}{c}{Pokec-n}                            \\
& \multicolumn{1}{l}{Acc.\%}   & \multicolumn{1}{l}{$\Delta_{\text{DP}}$\%}    & \multicolumn{1}{l||}{$\Delta_{\text{EO}}$\%}  
& \multicolumn{1}{l}{Acc.\%}   & \multicolumn{1}{l}{$\Delta_{\text{DP}}$\%}    & \multicolumn{1}{l||}{$\Delta_{\text{EO}}$\%}   
& \multicolumn{1}{l}{Acc.\%}   & \multicolumn{1}{l}{$\Delta_{\text{DP}}$\%}    & \multicolumn{1}{l}{$\Delta_{\text{EO}}$\%}   \\
\hline
\hline                          
FASD  & $\mathbf{69.22}_{(0.96)}$ & $\mathbf{0.92}_{(0.66)}$   & $\mathbf{4.47}_{(0.23)}$  & ${66.15}_{(0.49)}$ & $\mathbf{2.28}_{(0.42)}$  & $\mathbf{1.96}_{(0.51)}$  &  $\mathbf{66.34}_{(0.37)}$  & $\mathbf{0.79}_{(0.38)}$   & $\mathbf{0.91}_{(0.53)}$      \\
$\; -$ w/o Diffusion   & ${67.52}_{(2.17)}$ & ${3.29}_{(2.88)}$   & ${4.77}_{(1.03)}$ & ${67.71}_{(0.33)}$ & ${3.85}_{(0.15)}$  & ${4.71}_{(0.47)}$ &  ${65.02}_{(1.59)}$  & ${2.74}_{(1.25)}$   & ${2.84}_{(1.84)}$      \\
$\; -$ w/o Fairness & ${67.52}_{(1.70)}$ & ${3.10}_{(1.34)}$   & ${4.94}_{(4.25)}$ & $\mathbf{67.78}_{(0.45)}$ & ${4.81}_{(0.93)}$  & ${5.17}_{(1.10)}$ &  ${65.47}_{(0.71)}$  & ${1.74}_{(0.86)}$   & ${1.71}_{(0.81)}$      \\
\hline
\end{tabular} }
\label{table:ablation}
\vskip -0.20in
\end{table}

\subsection{Ablation Study}

We conducted an ablation study to investigate 
our proposed methodology, considering two variants: 
(1) Drop Fairness-Aware Subgraph Diffusion (w/o Diffusion): 
the node classification model $f$ is trained directly on the sampled subgraphs without incorporating Fairness-Aware Subgraph Diffusion. (2) Drop Fairness-Based Perturbations in the forward diffusion (w/o Fairness): 
only stochastic perturbations are applied in the forward diffusion process, thereby discarding the fairness-aware component of subgraph forward diffusion. 
The ablation study results are presented in Table \ref{table:ablation}.

Table \ref{table:ablation} illustrates the substantial improvement in fairness achieved 
by our full proposed method compared the two variants, across both equal opportunity and demographic parity metrics on all the three datasets. 
Furthermore, the full proposed method achieves the highest accuracy on both NBA and Pokec-n datasets. 
The Drop Fairness-Aware Subgraph Diffusion variant (w/o Diffusion) yields lower fairness results in both fairness evaluation metrics across all the three datasets, underscoring the tendency for sampled subgraphs to inherit biases present in the data. 
Similarly, the Drop Fairness-Based Perturbations variant (w/o Fairness) exhibits a decline in performance regarding fairness metrics on all the three datasets, highlighting the limitations of standard graph diffusion methods in enhancing fairness within the context of Fair GNN learning.

\begin{figure}[t]
\centering
\begin{subfigure}{0.32\textwidth}
\centering
\includegraphics[width = \textwidth]{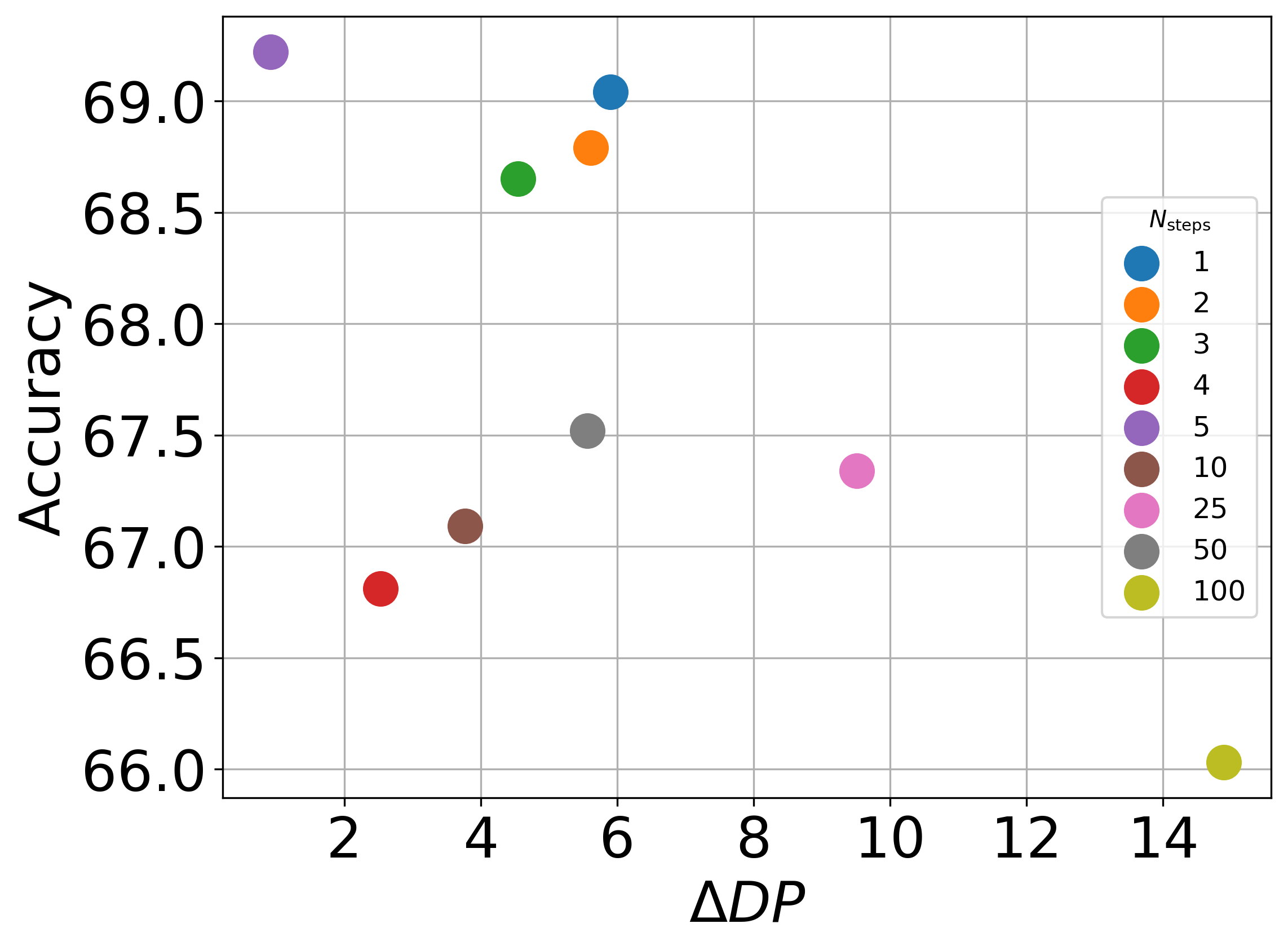} 
\caption{$N_{\text{steps}}$, NBA}
\label{fig:NBA_Acc_DP_alpha}
\end{subfigure}
\begin{subfigure}{0.32\textwidth}
\centering
\includegraphics[width = \textwidth]{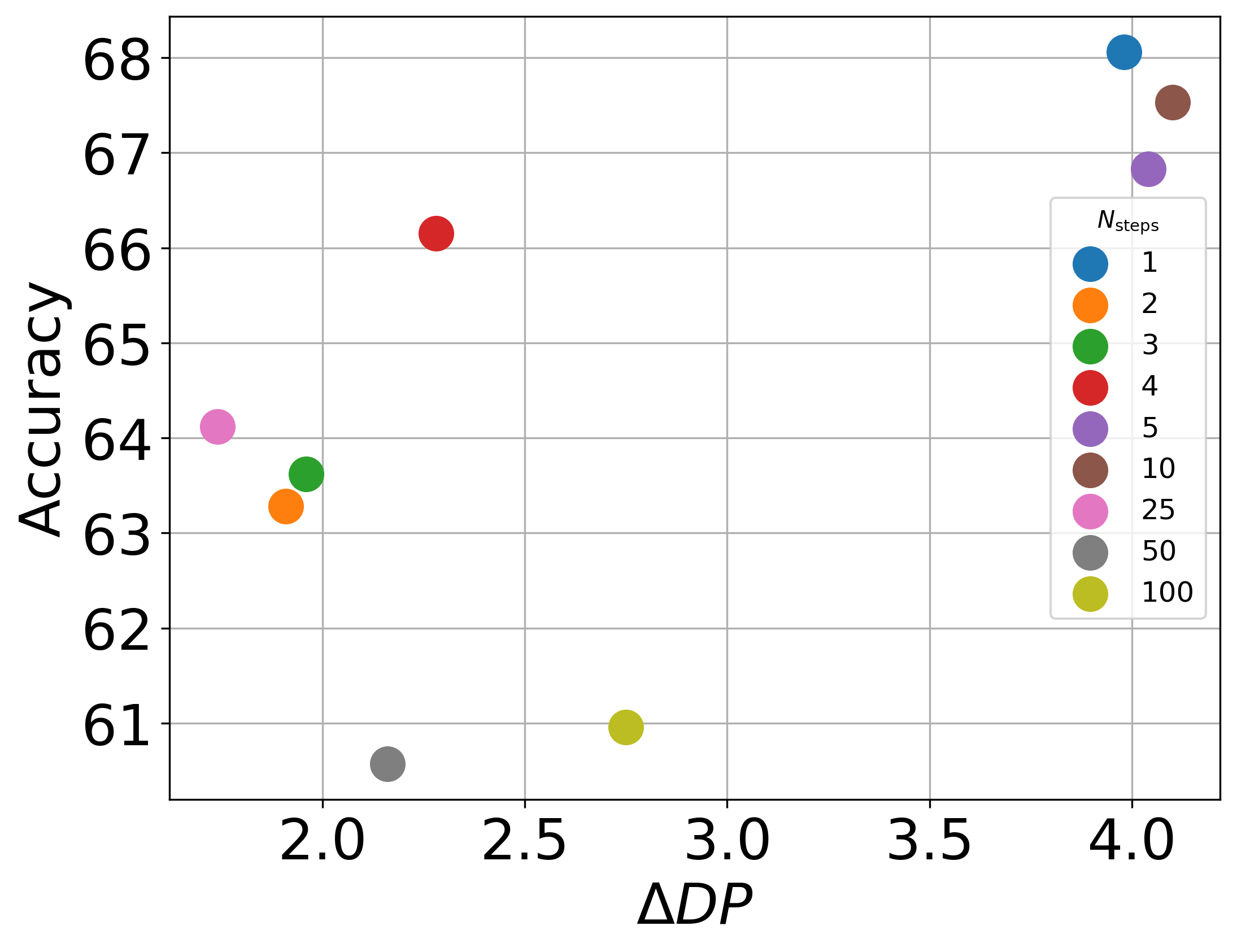}
\caption{$N_{\text{steps}}$, Pokec-z}
\label{fig:Pokec_Z_Acc_DP_alpha}
\end{subfigure}
\begin{subfigure}{0.32\textwidth}
\centering
\includegraphics[width = \textwidth]{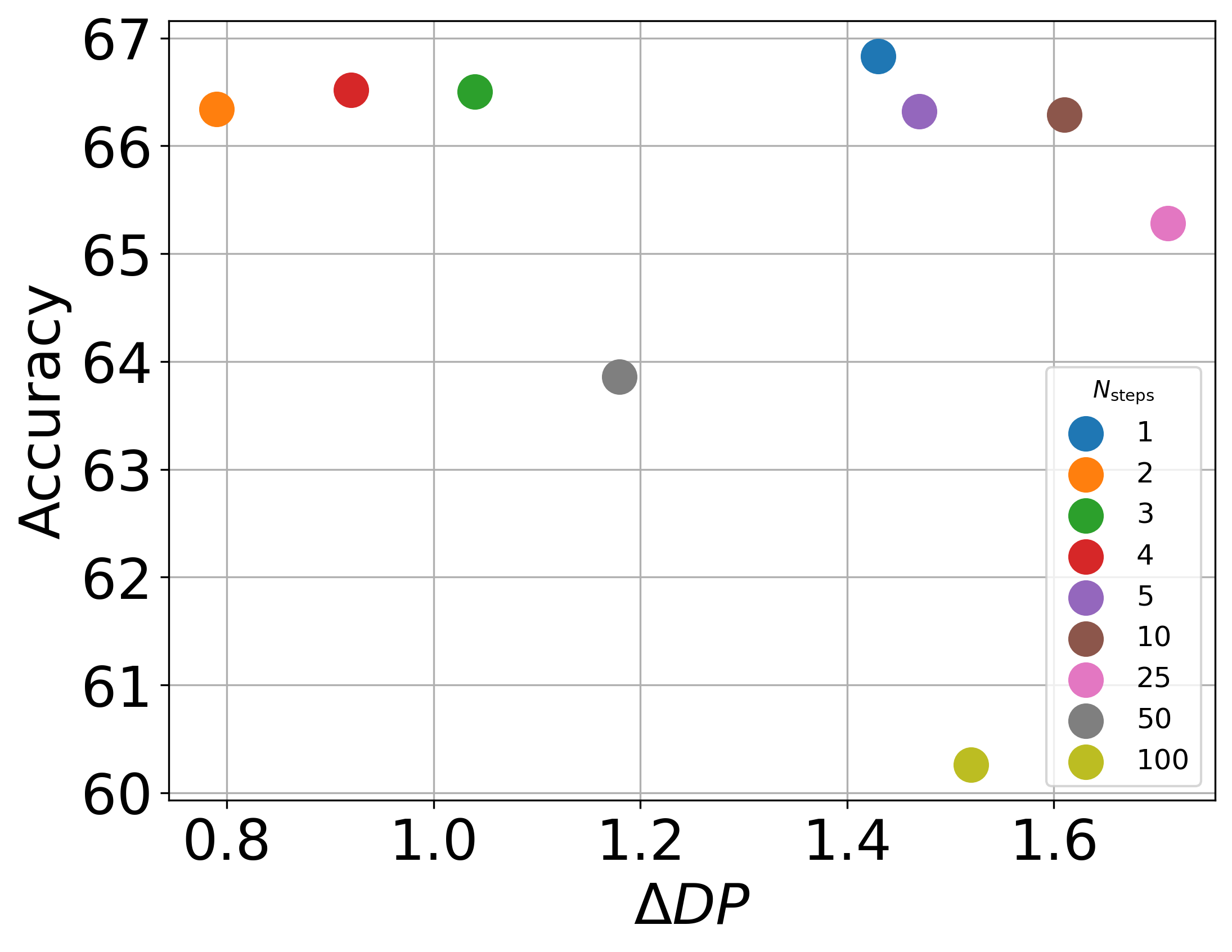}
\caption{$N_{\text{steps}}$, Pokec-n}
\label{fig:Pokec_N_Acc_DP_alpha}
\end{subfigure}

\begin{subfigure}{0.32\textwidth}
\centering
\includegraphics[width = \textwidth]{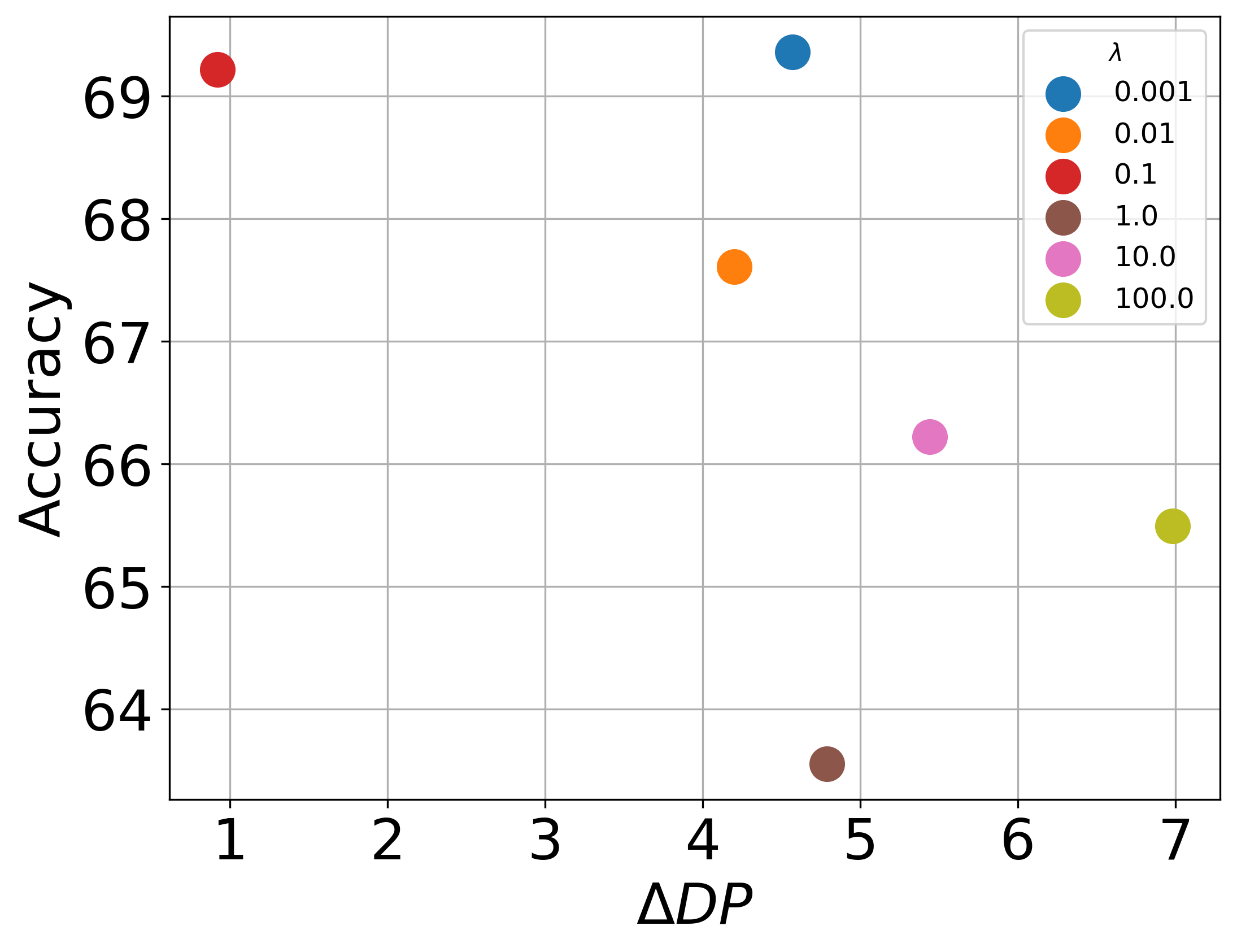} 
\caption{($\lambda_X$, $\lambda_A$), NBA}
\label{fig:NBA_Acc_DP_lambda}
\end{subfigure}
\begin{subfigure}{0.32\textwidth}
\centering
\includegraphics[width = \textwidth]{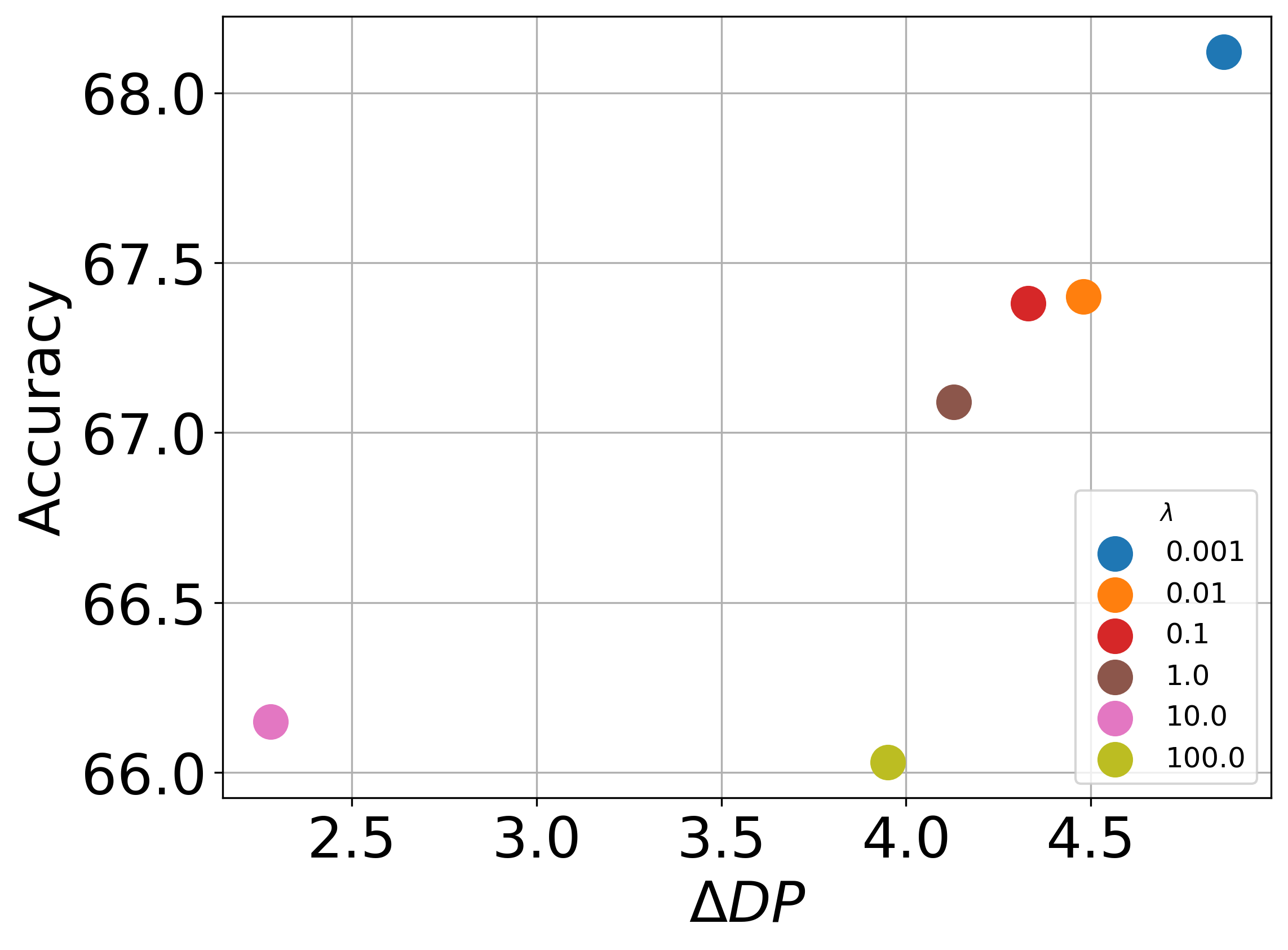}
\caption{($\lambda_X$, $\lambda_A$), Pokec-z}
\label{fig:Pokec_Z_Acc_DP_lambda}
\end{subfigure}
\begin{subfigure}{0.32\textwidth}
\centering
\includegraphics[width = \textwidth]{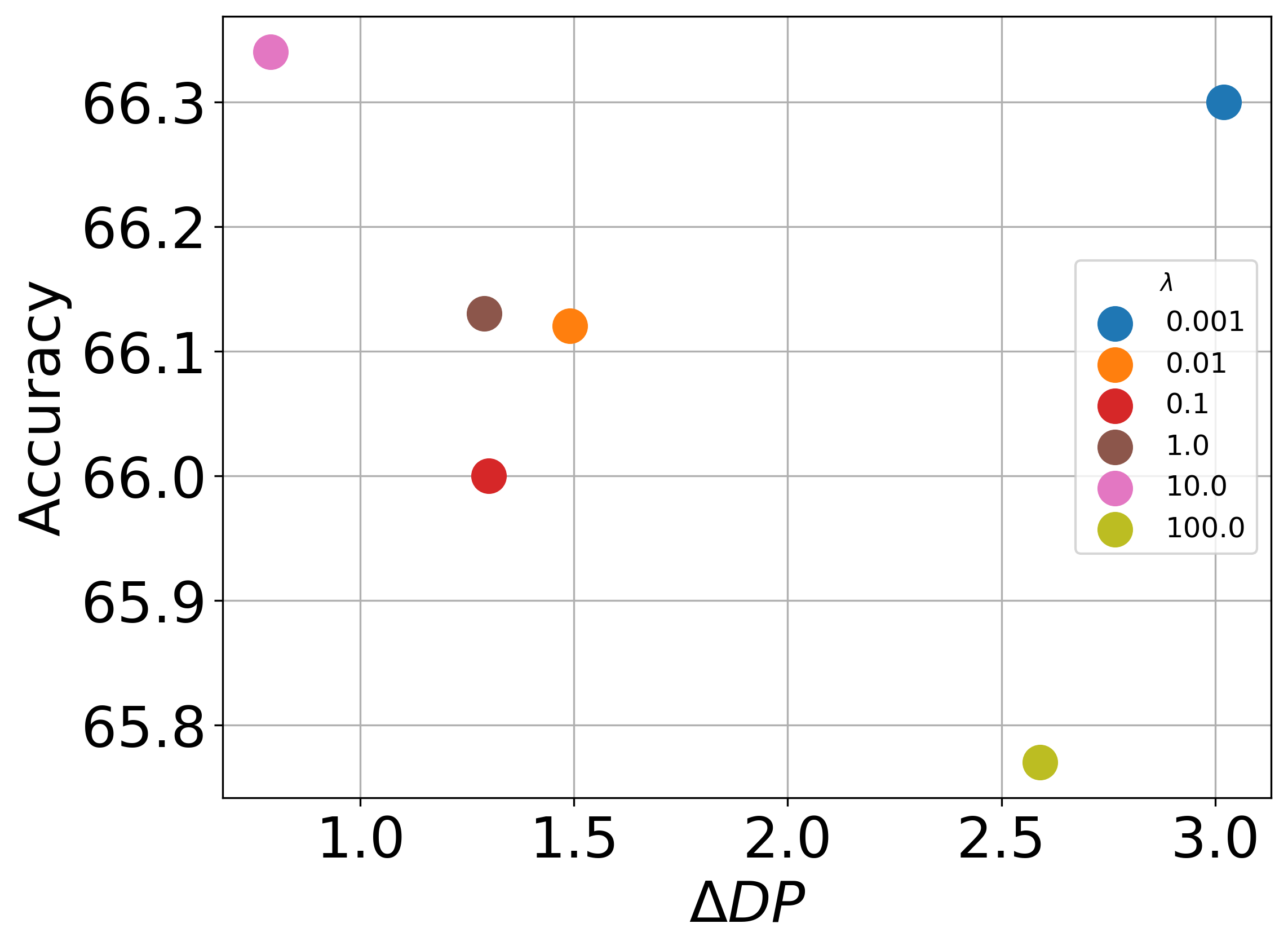}
\caption{($\lambda_X$, $\lambda_A$), Pokec-n}
\label{fig:Pokec_N_Acc_DP_lambda}
\end{subfigure}
\caption{Sensitivity analysis for our proposed FASD method in terms of Accuracy Vs. $\Delta \text{DP}$ on hyper-parameters $N_{\text{steps}}$ and ($\lambda_X$, $\lambda_A$): (a) $N_{\text{steps}}$, NBA; (b) $N_{\text{steps}}$, Pokec-z; (c) $N_{\text{steps}}$, Pokec-n; (d) ($\lambda_X$, $\lambda_A$), NBA; (e) ($\lambda_X$, $\lambda_A$), Pokec-z; (f) ($\lambda_X$, $\lambda_A$), Pokec-n.}

\label{fig:hyper_sen}
\vskip -0.15in

\end{figure} 

\subsection{Hyper-Parameter Sensitivity}

To investigate the impact of the hyper-parameters in our proposed FASD method, we present the results of several sensitivity analyses on NBA, Pokec-z and Pokec-n datasets in Figure \ref{fig:hyper_sen}. Specifically, Figures \ref{fig:NBA_Acc_DP_alpha}, \ref{fig:Pokec_Z_Acc_DP_alpha}, \ref{fig:Pokec_N_Acc_DP_alpha} illustrate the impact of varying the number of reverse-diffusion steps $N_{\text{steps}}$ on both accuracy and demographic parity. 
Additionally, Figures \ref{fig:NBA_Acc_DP_lambda}, \ref{fig:Pokec_Z_Acc_DP_lambda}, and \ref{fig:Pokec_N_Acc_DP_lambda} depict the effects of adjusting the hyperparameters $\lambda_X$ and $\lambda_A$, which control the contribution of fairness-based perturbations in the fairness-aware forward diffusion process.

In the case of $N_{\text{steps}}$ hyper-parameter, the top-left corner point represents the $N_{\text{steps}}$ value that optimally balances high accuracy with low bias (high fairness). 
Across all datasets, our findings indicate that optimal performance is achieved with a relatively small number of reverse-diffusion steps ($\{2,3,4,5\}$). Excessive steps ($N_{\text{steps}} > 5$) result in diminished model accuracy and fairness, while a single step ($N_{\text{steps}}=1$) yields high accuracy but with pronounced bias. This underscores the necessity of multiple reverse-diffusion steps to effectively debias the input subgraphs. 
Similarly, for $\lambda_X$ and $\lambda_A$ hyper-parameters, the top-left corner point represents the $\lambda_X$ and $\lambda_A$ values that strike the ideal balance between accuracy and fairness. 
Our results reveal that very small values ($\lambda_X$, $\lambda_A$ < $0.1$) offer marginal improvement in fairness, while excessively large values ($\lambda_X$, $\lambda_A$ > $10$) significantly impair accuracy. Values within the range of $[0.1,10.0]$ achieve the optimal balance, with $0.1$ yielding the best results for NBA dataset and $10.0$ yielding optimal performance for Pokec-z and Pokec-n.

\section{Conclusion}
In this paper, we introduced a novel Fairness-Aware Subgraph Diffusion (FASD) method 
to tackle fair GNN learning for graph node classification. 
FASD generates fairness-based subgraph perturbations using an adversary sensitive attribute prediction model. 
These perturbations are then integrated into the fairness-aware forward diffusion process. 
Through the training of score-based models to predict the applied perturbation scores, 
we enable the learning of the underlying dynamics of fairness and bias within the subgraph data. 
Subsequently, the trained score-based models are employed to debias the subgraph samples via 
the reverse-diffusion process. 
Finally, standard node classification models are trained on the debiased subgraphs to yield 
fair and accurate predictions. 
Experimental evaluations conducted on three benchmark datasets validate 
the effectiveness of our proposed method. 
The empirical results demonstrate the capacity of the proposed FASD
for successfully mitigating bias inherent in the data, outperforming state-of-the-art Fair GNN methods.

\bibliographystyle{abbrvnat}
\bibliography{main}

\newpage

\appendix


\section{Subgraph Sampling Procedure}
\label{section:subgraph_sampling}

The details of the subgraph sampling procedure are presented in Algorithm \ref{alg:subgraph_sampling}.

\begin{algorithm}[h]
    \caption{Subgraph Sampling Procedure}
    \label{alg:subgraph_sampling}
    \begin{algorithmic}[1]
\STATE{\textbf{Input:} Graph $G(V,E)$, Starting node $u$, Hyper-parameters $d$, $k$}
\STATE{\textbf{Output:} Subgraph $G^{(i)}(V^{(i)},E^{(i)})$}
\STATE Initialize $Q \leftarrow [u]$
\STATE Initialize $V^{(i)} \leftarrow \{u\}$
\STATE Initialize $E^{(i)} \leftarrow \{\}$
\STATE Initialize $visited \leftarrow \{\}$
\WHILE{$Q$ is not empty \textbf{and} $d > 0$}
    \STATE $v \leftarrow$ Dequeue from $Q$
    \IF{$v$ is not in $visited$}
        \STATE Add $v$ to $visited$
        \STATE $\mathcal{N}_v$ $\leftarrow$ Sample $k$ neighbors of $v$  
        \STATE Append $\mathcal{N}_v$ to $V^{(i)}$
        \STATE Add edges between $v$ and $\mathcal{N}_v$ to $E^{(i)}$
        \STATE Enqueue $\mathcal{N}_v$ into $Q$
        \STATE Decrement $d$ by $1$
    \ENDIF
\ENDWHILE
\STATE \textbf{return} $G^{(i)}(V^{(i)},E^{(i)})$ 
\end{algorithmic}
\end{algorithm}

\section{Perturbation Kernels}
\label{section:kernels}

We utilize the perturbation kernels $\psi_{1,t}(.)$ and $\psi_{2,t}(.)$ provided in \cite{song2020score}. 
The mean and standard deviation of $X^{(i)}_0$ at time $t$ are calculated using perturbation kernel $\psi_{1,t}(.)$ as follows:
\begin{equation}
\begin{aligned}
& \mu_{t}(X^{(i)}) = \text{exp}\bigl(\frac{-t^2(\beta_{1,\text{max}} - \beta_{1,\text{min}})}{4} - \frac{t \beta_{1,\text{min}} }{2}   \bigr)  X^{(i)} \\
& \sigma_{t}(X^{(i)}) = \sqrt{1 - \text{exp}\bigl( 2 (\frac{-t^2(\beta_{1,\text{max}} - \beta_{1,\text{min}})}{4}   -\frac{t \beta_{1,\text{min}} }{2} )  \bigr)}
\end{aligned}
\label{eq:kernel}
\end{equation}
where $\beta_{1,\text{max}}$ and $\beta_{1,\text{min}}$ are hyper-parameters of perturbation kernel $\psi_{1,t}(.)$ that govern the rate at which perturbations are applied to the data over time, and $\text{exp(.)}$ is the exponential function. 
Similarly, the mean and standard deviation of $A^{(i)}_0$ at time $t$ are calculated using perturbation kernel $\psi_{2,t}(.)$ as follows:
\begin{equation}
\begin{aligned}
& \mu_{t}(A^{(i)}) = \text{exp}\bigl(\frac{-t^2(\beta_{2,\text{max}} - \beta_{2,\text{min}})}{4} - \frac{t \beta_{2,\text{min}} }{2}   \bigr)  A^{(i)} \\
& \sigma_{t}(A^{(i)}) = \sqrt{1 - \text{exp}\bigl( 2 (\frac{-t^2(\beta_{2,\text{max}} - \beta_{2,\text{min}})}{4}   -\frac{t \beta_{2,\text{min}} }{2} )  \bigr)}
\end{aligned}
\label{eq:kernel_2}
\end{equation}
where $\beta_{2,\text{max}}$ and $\beta_{2,\text{min}}$ are hyper-parameters of perturbation kernel $\psi_{2,t}(.)$. 
The standard deviations $\sigma_{t}(X^{(i)})$ and $\sigma_{t}(A^{(i)})$ depend solely on time $t$ and increase asymptotically towards $\sigma_{t} = \sigma_{t}(X^{(i)}) = \sigma_{t}(A^{(i)}) = 1$ as $t$ increases. Conversely, the means $\mu_{t}(X^{(i)})$ and $\mu_{t}(A^{(i)})$ decrease as $t$ increases.

\section{Score-based Models Training Procedure}
\label{section:score_training}

The details of the training procedure of the score-based models are provided in Algorithm \ref{alg:score_train}.
    
\begin{algorithm}[h]
    \caption{Score-Based Models Training Procedure}
    \label{alg:score_train}
    \begin{algorithmic}[1]
\STATE{\textbf{Input:} Subraph set $\mathcal{G}$, trained sensititive attribute predictor $g_{\text{sen}}$,\\ Perturbation kernels $\psi_{1,t}(.)$ and $\psi_{2,t}(.)$, Hyper-parameter $\lambda_X$, $\lambda_A$, $T$}
\STATE{\textbf{Output:} Learned parameters of score-based models  $\theta$, $\phi$}
\FOR{{iter = 1} {\bf to} maxiters}
        \STATE Sample $t$ from $[0,T]$
        \STATE Sample $G^{(i)}_0(X^{(i)}_0,A^{(i)}_0)$ from $\mathcal{G}$
    \STATE Sample $\epsilon_X \sim \mathcal{N}(0, 1)^{(N^{(i)}, D)}$
    \STATE Sample $\epsilon_A \sim \mathcal{N}(0, 1)^{(N^{(i)}, N^{(i)})}$
    \STATE Calculate $\mu_{t}(X^{(i)}_0), \, \sigma_{t}(X^{(i)}_0) = \psi_{1,t}(X^{(i)}_0)$ using Eq.(\ref{eq:kernel})
    \STATE Calculate $\mu_{t}(A^{(i)}_0), \, \sigma_{t}(A^{(i)}_0) = \psi_{2,t}(A^{(i)}_0)$ using Eq.(\ref{eq:kernel_2})
    \STATE Calculate $\nabla_{X}\mathcal{L}_{\text{sen}}(X^{(i)}_0,A^{(i)}_0)$, $\gamma_X$ using Eq.(\ref{eq:g_sen}), Eq.(\ref{eq:sen_att_loss}), Eq.(\ref{eq:gamma})
    \STATE Calculate $\nabla_{A}\mathcal{L}_{\text{sen}}(X^{(i)}_0,A^{(i)}_0)$, $\gamma_A$  using Eq.(\ref{eq:g_sen}), Eq.(\ref{eq:sen_att_loss}), Eq.(\ref{eq:gamma})
    \STATE Apply Fairness-Aware Forward Diffusion 
    using Eq.(\ref{eq:forward_sub_pert})
    \STATE Generate predictions of score-based models $s_{\theta,t}(G^{(i)}_t)$, $s_{\phi,t}(G^{(i)}_t)$ using Eq.(\ref{eq:s_theta}), Eq. (\ref{eq:s_phi})
    \STATE Calculate loss $\mathcal{L}_{\theta}(t,G^{(i)}_0,G^{(i)}_t) = || s_{\theta,t}(G^{(i)}_t) - \epsilon_X + \frac{\gamma_X}{\sigma_t(X^{(i)}_0)}   \nabla_{X}\mathcal{L}_{\text{sen}}(X^{(i)}_0,A^{(i)}_0)||_2^2$
    \STATE Calculate loss $\mathcal{L}_{\phi}(t,G^{(i)}_0,G^{(i)}_t) =  || s_{\phi,t}(G^{(i)}_t) -  \epsilon_A   + \frac{\gamma_A}{\sigma_t(A^{(i)}_0)}   \nabla_{A}\mathcal{L}_{\text{sen}}(X^{(i)}_0,A^{(i)}_0)||_2^2$
    \STATE Update $\theta$ to minimize $\mathcal{L}_{\theta}(t,G^{(i)}_0,G^{(i)}_t)$ with gradient descent
    \STATE Update $\phi$ to minimize $\mathcal{L}_{\phi}(t,G^{(i)}_0,G^{(i)}_t)$ with gradient descent
    
\ENDFOR
\STATE \textbf{return} $\theta$, $\phi$
\end{algorithmic}
\end{algorithm}

\section{PC Sampler}
\label{section:sampler}

We adopt Euler-Maruyama numerical solver \cite{Kloeden1992} for the two Predictors ($\text{Predictor}_{X}$, $\text{Predictor}_{A}$) and Langevin MCMC \cite{parisi1981correlation} for the two Correctors ($\text{Corrector}_{X}$, $\text{Corrector}_{A}$). 
In the case of the Euler-Maruyama Predictor, we utilize the implementation provided in \cite{jo2022score} where at each reverse-diffusion step ($j$) the debiased estimates of $X^{(i)}$ and $A^{(i)}$ are updated as follows: 
\begin{equation}
\begin{aligned}
    & \quad \quad \quad \quad \quad \quad  \quad \quad \beta_{j+1} = \beta_{\text{min}} + (j+1)  (\beta_{\text{max}}- \beta_{\text{min}}) \\
    &  X_{j}^{(i)} = X_{j+1}^{(i)} + \frac{1}{2} \beta_{j+1}   X_{j+1}^{(i)}  + \beta_{j+1} s_{\theta,j+1}(G_{j+1}^{(i)}) +\sqrt{\beta_{j+1}}  \epsilon_X  \\ 
    &  A_{j}^{(i)} = A_{j+1}^{(i)} + \frac{1}{2} \beta_{j+1}   A_{j+1}^{(i)}  + \beta_{j+1} s_{\phi,j+1}(G_{j+1}^{(i)}) +\sqrt{\beta_{j+1}}  \epsilon_A 
\end{aligned}
\label{eq:predictor}
\end{equation}
where $\beta_{\text{min}}$ and $\beta_{\text{max}}$ are hyper-parameters of perturbation kernels $\psi_{1,t}(.)$ and $\psi_{2,t}(.)$. $\epsilon_X$ and $\epsilon_A$ are sampled from $\mathcal{N}(0, 1)^{(N^{(i)}, D)}$ and $\mathcal{N}(0, 1)^{(N^{(i)}, N^{(i)})}$, respectively. $G_{j+1}^{(i)}$ is made up of node input features $X_{j+1}^{(i)}$ and subgraph adjacency matrix $A_{j+1}^{(i)}$. 
As for the Langevin MCMC Correctors, we adopt the implementation provided in \cite{song2020score} where at each reverse-diffusion step ($j$) the debiased estimates of $X^{(i)}$ and $A^{(i)}$ are updated as follows: 
\begin{equation}
\begin{aligned}
    &        X_{j}^{(i)} = X_{j+1}^{(i)} + \omega_X  s_{\theta,j+1}( G_{j+1}^{(i)}) +  \sqrt{2 \omega_X}  \epsilon_X \\
    &  A_{j}^{(i)} = A_{j+1}^{(i)} + \omega_A  s_{\phi,j+1}( G_{j+1}^{(i)}) +  \sqrt{2 \omega_A}  \epsilon_A   \\ 
    & \quad \quad \quad  \omega_X = 2  \left( \frac{r_X ||\epsilon_X ||_2}{||s_{\theta,j+1}( G_{j+1}^{(i)}) ||_2} \right)^2\\
    & \quad \quad \quad  \omega_A = 2  \left( \frac{r_A ||\epsilon_A ||_2}{||s_{\phi,j+1}( G_{j+1}^{(i)}) ||_2} \right)^2
\end{aligned}
\label{eq:corrector}
\end{equation}

where $r_X$ and $r_A$ are hyper-parameters controlling the signal-to-noise ratio for $X^{(i)}$ and $A^{(i)}$ respectively. $\epsilon_X$ and $\epsilon_A$ are sampled from $\mathcal{N}(0, 1)^{(N^{(i)}, D)}$ and $\mathcal{N}(0, 1)^{(N^{(i)}, N^{(i)})}$, respectively. The details of the PC sampler approximating the reverse-diffusion debiasing procedure are presented in Algorithm \ref{alg:pc_sampling}. 
    
\begin{algorithm}[h]
    \caption{PC Sampler Reverse-Diffusion Debiasing Procedure}
    \label{alg:pc_sampling}
    \begin{algorithmic}[1]
\STATE{\textbf{Input:} Subraph $G^{(i)}(X^{(i)},A^{(i)})$, Trained score-based models $s_{\theta,t}$,  $s_{\phi,t}$,\\   Hyper-parameter $N_{\text{steps}}$}
\STATE{\textbf{Output:} Debiased Subgraph $\tilde{G}^{(i)}(\tilde{X}^{(i)},\tilde{A}^{(i)})$}
\STATE Initialize $X_{N_{\text{steps}}}^{(i)} \leftarrow X^{(i)}$, $A_{N_{\text{steps}}}^{(i)} \leftarrow A^{(i)}$.

\FOR{iter = $N_{\text{steps}}-1$ to 0}
    \STATE $\hat{X}_{\text{iter}}^{(i)} = \text{Predictor}_{X}(X_{\text{iter}+1}^{(i)},A_{\text{iter}+1}^{(i)},s_{\theta,\text{iter}+1})$ using Eq.(\ref{eq:predictor})
    \STATE $\hat{A}_{\text{iter}}^{(i)} = \text{Predictor}_{A}(X_{\text{iter}+1}^{(i)},A_{\text{iter}+1}^{(i)},s_{\phi,\text{iter}+1})$ using Eq.(\ref{eq:predictor})
    \STATE $X_{\text{iter}}^{(i)} = \text{Corrector}_{X}(\hat{X}_{\text{iter}}^{(i)},\hat{A}_{\text{iter}}^{(i)},s_{\theta,\text{iter}+1})$ using Eq.(\ref{eq:corrector})
    \STATE $A_{\text{iter}}^{(i)} = \text{Corrector}_{A}(\hat{X}_{\text{iter}}^{(i)},\hat{A}_{\text{iter}}^{(i)},s_{\phi,\text{iter}+1})$ using Eq.(\ref{eq:corrector})
\ENDFOR
\STATE Assign $\tilde{X}^{(i)} \leftarrow X_{0}^{(i)}$, $\tilde{A}^{(i)} \leftarrow A_{0}^{(i)}$.
\STATE \textbf{return} $\tilde{G}^{(i)}(\tilde{X}^{(i)},\tilde{A}^{(i)})$
\end{algorithmic}
\end{algorithm}

\section{Benchmark Datasets}
\label{section:dataset}
In the NBA dataset, nodes represent NBA players, with node features comprising player information and performance statistics. The sensitive attribute pertains to nationality, while labels indicate whether a player's salary surpasses the median. 
Regarding the Pokec-z and Pokec-n datasets, both are derived from the Pokec dataset, a prominent social network in Slovakia. Each node corresponds to a user profile, with features encompassing profile details such as hobbies, interests, and age. The sensitive attribute denotes the user's region, while labels reflect the user's field of work. 
For the train/validation/test splits, we allocate 20\%/35\%/45\%, 10\%/10\%/80\%, and 10\%/10\%/80\% of the graph nodes in the NBA, Pokec-z, and Pokec-n datasets, respectively.

\section{Implementation Details}
\label{section:implementation}

\paragraph{Subgraph-level Instance Sampling: }
For each dataset, we sample a number of subgraph instances ($M$) equal to the total number of train, validation, and test nodes 
in the graph, where each one of these nodes serves as the starting point for one subgraph instance.
The hyper-parameter $d$, which signifies the depth of the sampled subgraphs, is set to $2$ for the NBA dataset and $3$ for the Pokec-z and Pokec-N datasets. 
The hyper-parameter $k$, which determines the number of neighbors to sample, is set to $10$ for all three datasets.
\paragraph{Fairness-Aware Forward Diffusion: }
The sensitive attribute prediction model consists of two GCN layers \cite{kipf2017semisupervised} followed by two fully connected layers. Each GCN and fully connected layer is followed by a Rectified Linear Unit (ReLU) activation function. The hidden layer embeddings have sizes $(64, 32, 16)$, and dropout is applied to each layer with a dropout rate of $0.1$. We train the sensitive attribute prediction model for 500 epochs using the Adam optimizer with a learning rate of $1 \times 10^{-4}$. 
The hyper-parameters ($\lambda_X$, $\lambda_A$) are set to $0.1$ for the NBA dataset and $10$ for the Pokec-z and Pokec-n datasets. The hyper-parameters ($\beta_{1,\text{max}}$, $\beta_{2,\text{max}}$) and ($\beta_{1,\text{min}}$, $\beta_{2,\text{min}}$) are set to $1.0$ and $0.1$, respectively.
\paragraph{Estimating Diffusion Perturbations via Score-Based Models: }
In the case of the score-based model $s_{\theta,t}$, $\text{GNN}_{X}$ consists of 3 GCN layers followed by hyperbolic tangent (tanh) activation functions, while $\text{MLP}_{X}$ comprises 3 fully connected layers with ReLU activation. 
As for the score-based model $s_{\phi,t}$, $\text{GNN}_{X}$ is made up of 5 GCN layers, $\text{GMH}$ employs 4 attention heads with $P$ and $K$ taking values $2$ and $5$, respectively, and $\text{MLP}_{A}$ comprises 3 fully connected layers. Both the GCN layers and fully connected layers use the Exponential Linear Unit (ELU) activation function. 
The hidden embedding size across all layers is set to 32. 
The score-based models are trained jointly for 1000 epochs using the Adam optimizer with a learning rate of $1 \times 10^{-2}$ and weight decay of $1 \times 10^{-4}$. 
\paragraph{Subgraph Debiasing via Reverse-Diffusion: }
The number of steps of the reverse-diffusion process ($N_{\text{steps}}$) is set to 5, 4 and 2 for the NBA, Pokec-z and Pokec-n datasets, respectively. The threshold for pruning weak edges 
($\tau$) takes value $0.5$ across all three datasets. 
Hyper-parameters ($r_{X}$, $r_{A}$) take value $0.05$. 
\paragraph{Fair Node Classification with Debiased Subgraphs: }
The node classification model $f$ consists of 2 GCN layers followed by 2 fully connected layers. Each GCN layer and fully connected layer is followed by ReLU activation function. The hidden embedding size of all layers is $64$, and dropout is applied to all layers with a dropout rate of $0.3$. The model is trained for 500 epochs using the Adam optimizer with a learning rate of $1 \times 10^{-3}$ and weight decay of $1 \times 10^{-4}$. 

For the results of comparison methods, we refer to the outcomes from \cite{ling2022learning}. Our experiments are conducted on GeForce RTX 4090 and GeForce RTX 2080 Ti GPUs.

\begin{figure}[t]
\centering
\begin{subfigure}{0.32\textwidth}
\centering
\includegraphics[width = \textwidth]{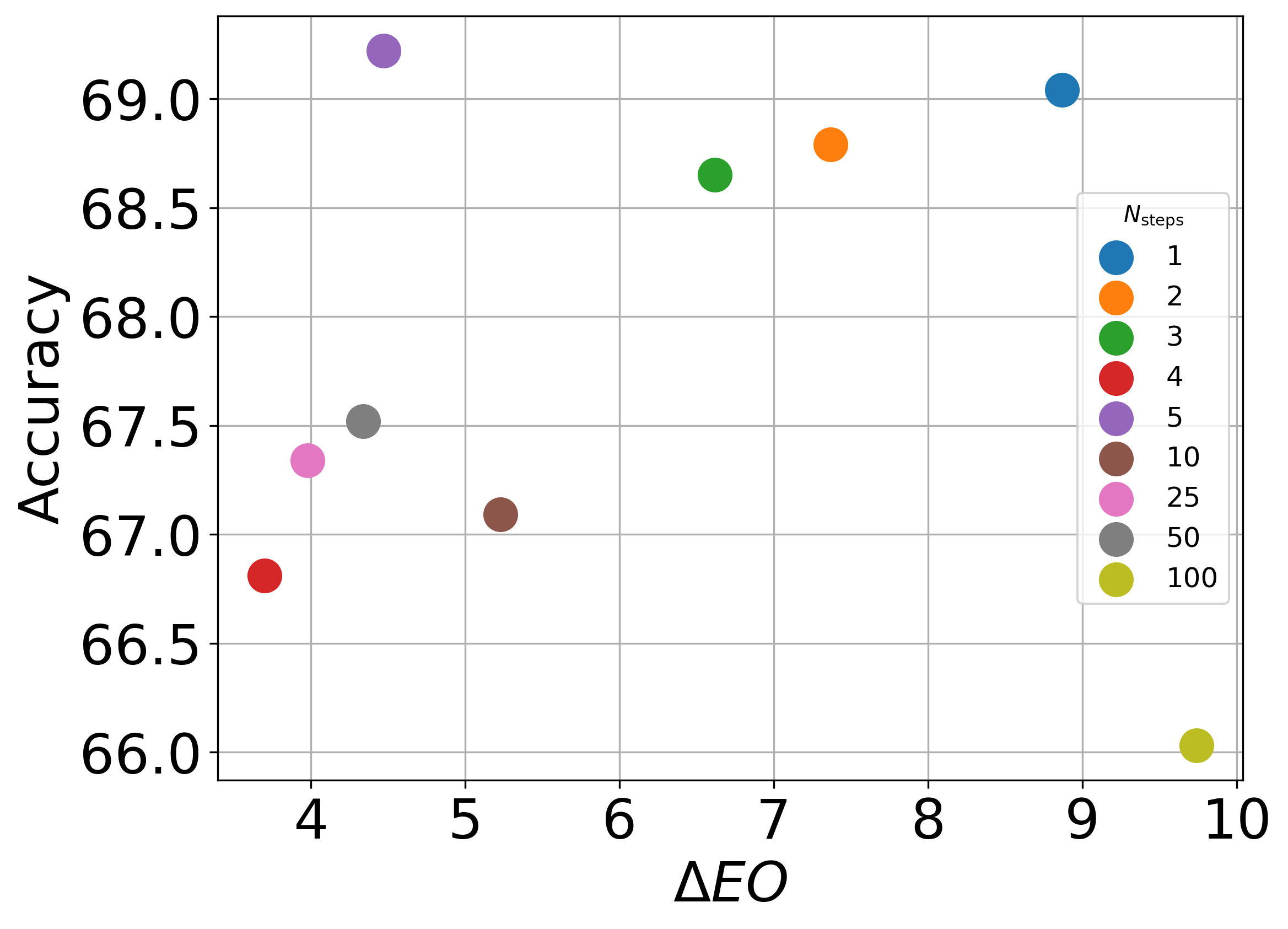} 
\caption{$N_{\text{steps}}$, NBA}
\label{fig:NBA_Acc_EO_alpha}
\end{subfigure}
\begin{subfigure}{0.32\textwidth}
\centering
\includegraphics[width = \textwidth]{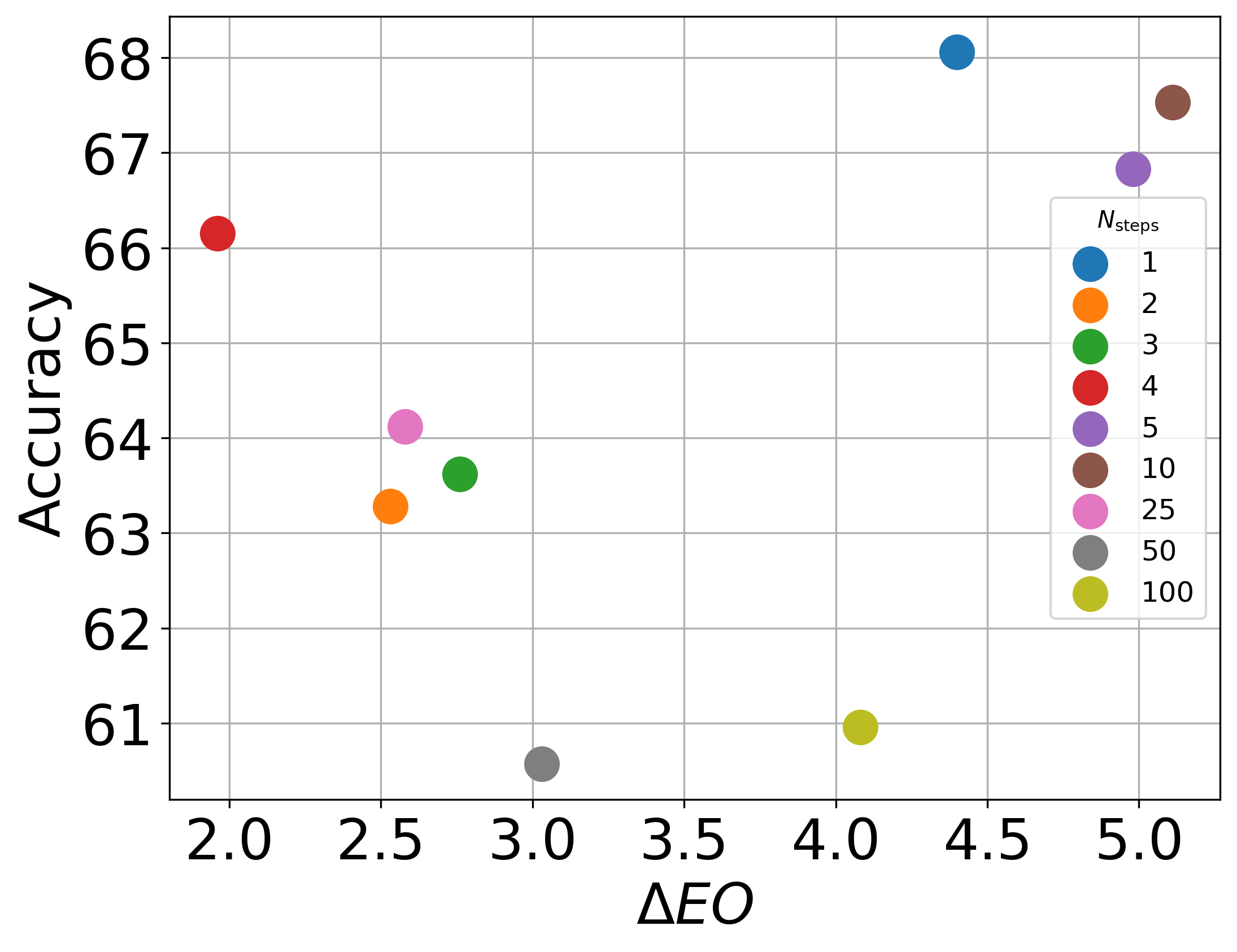}
\caption{$N_{\text{steps}}$, Pokec-z}
\label{fig:Pokec_Z_Acc_EO_alpha}
\end{subfigure}
\begin{subfigure}{0.32\textwidth}
\centering
\includegraphics[width = \textwidth]{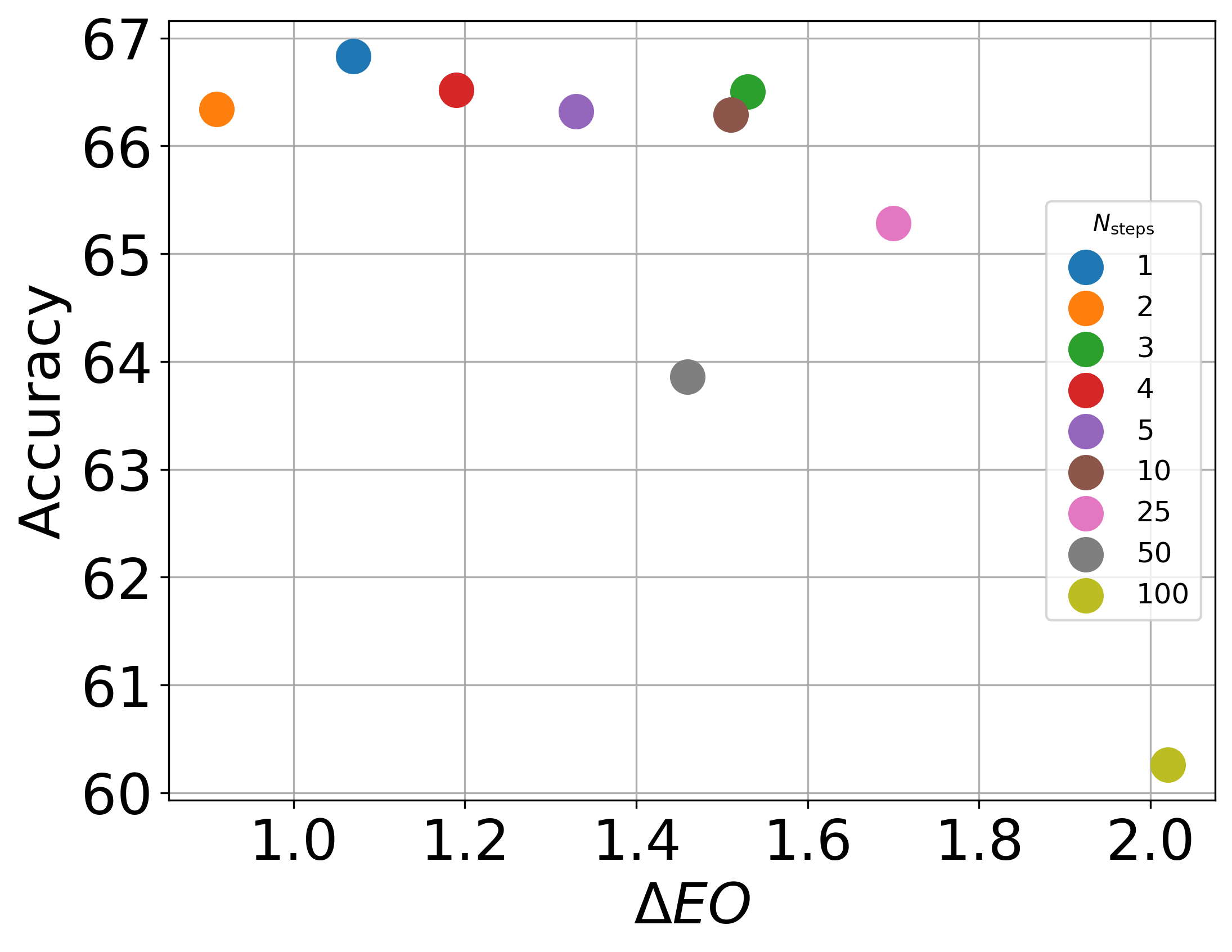}
\caption{$N_{\text{steps}}$, Pokec-n}
\label{fig:Pokec_N_Acc_EO_alpha}
\end{subfigure}
\begin{subfigure}{0.32\textwidth}
\centering
\includegraphics[width = \textwidth]{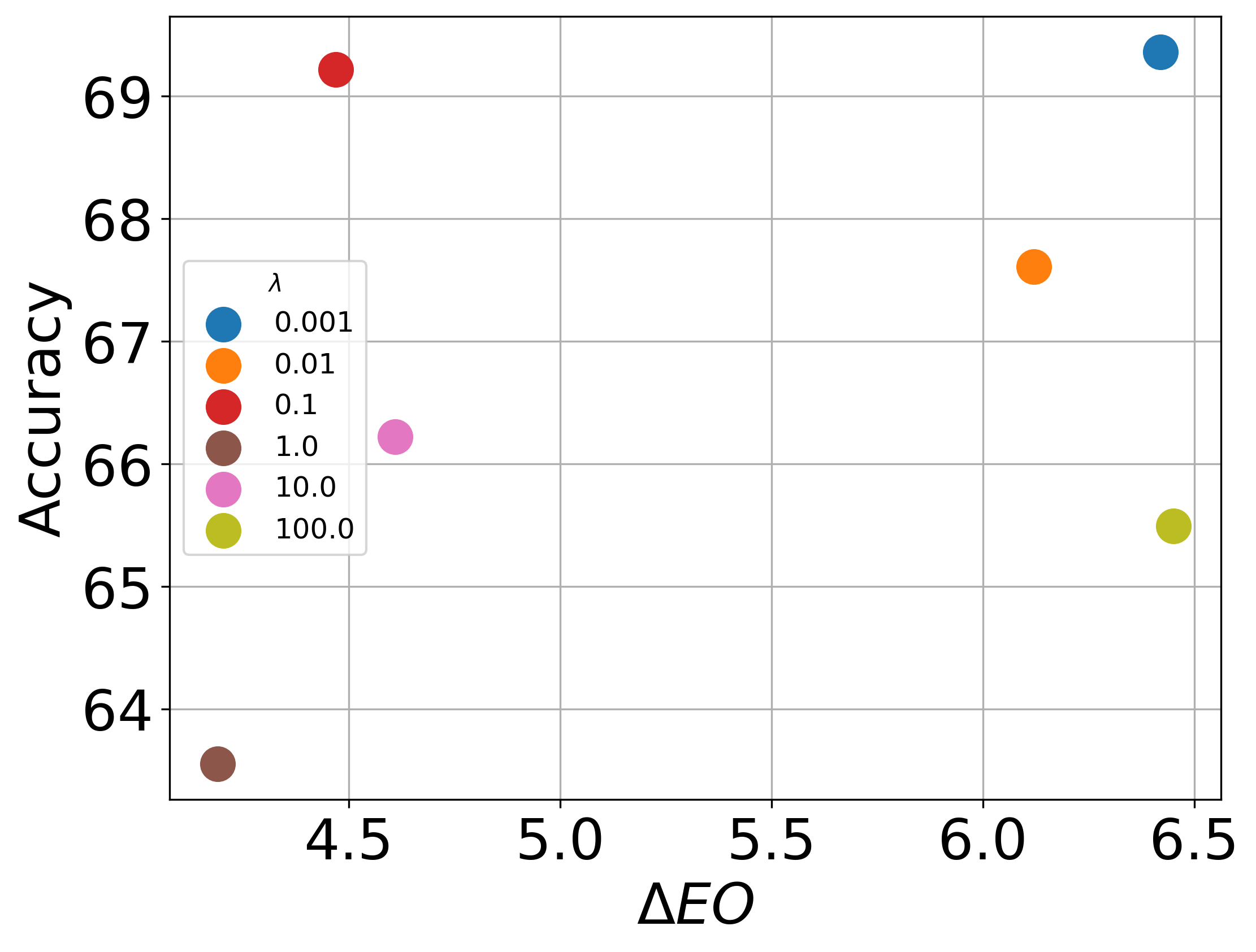} 
\caption{($\lambda_X$, $\lambda_A$), NBA}
\label{fig:NBA_Acc_EO_lambda}
\end{subfigure}
\begin{subfigure}{0.32\textwidth}
\centering
\includegraphics[width = \textwidth]{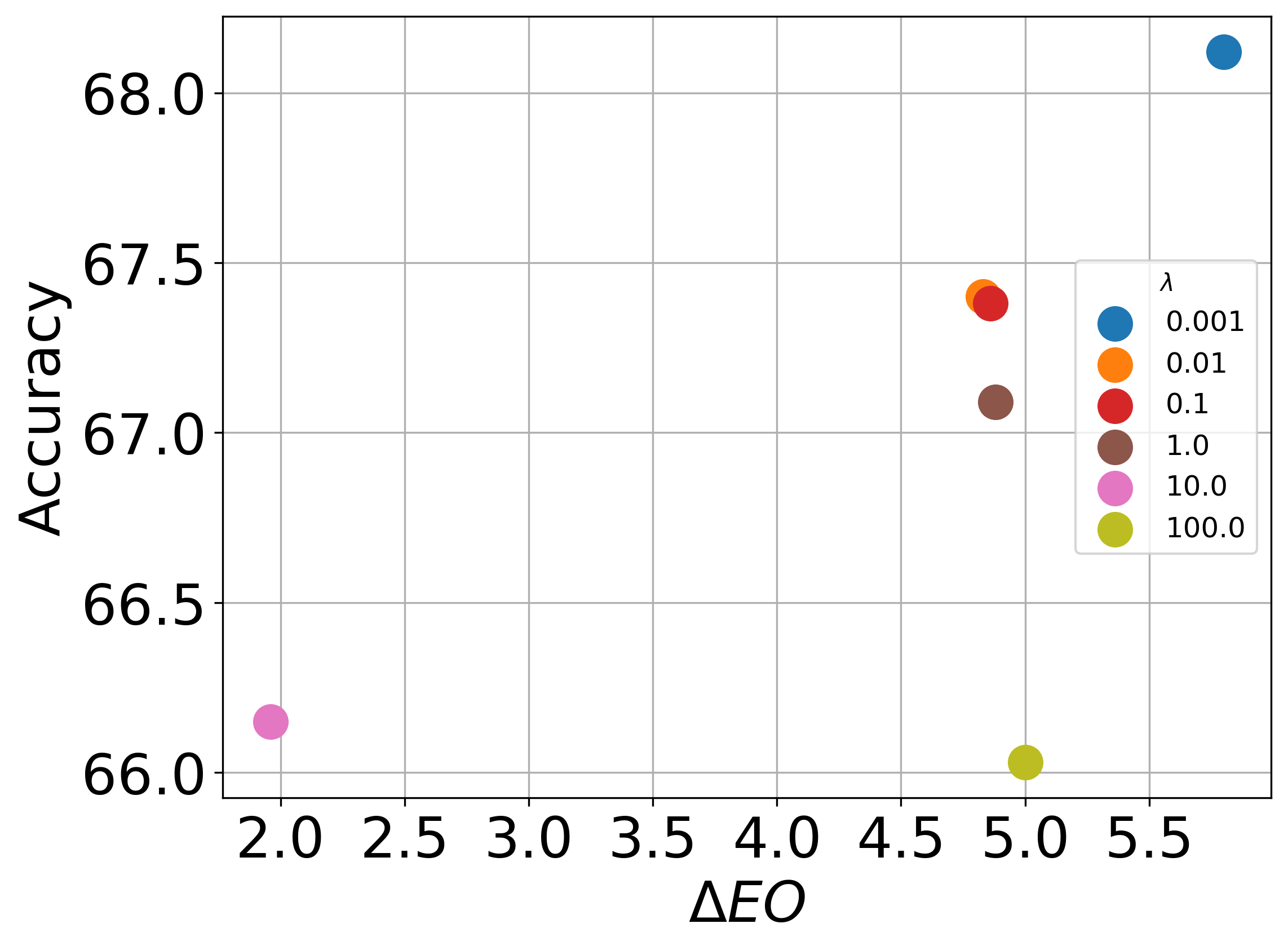}
\caption{($\lambda_X$, $\lambda_A$), Pokec-z}
\label{fig:Pokec_Z_Acc_EO_lambda}
\end{subfigure}
\begin{subfigure}{0.32\textwidth}
\centering
\includegraphics[width = \textwidth]{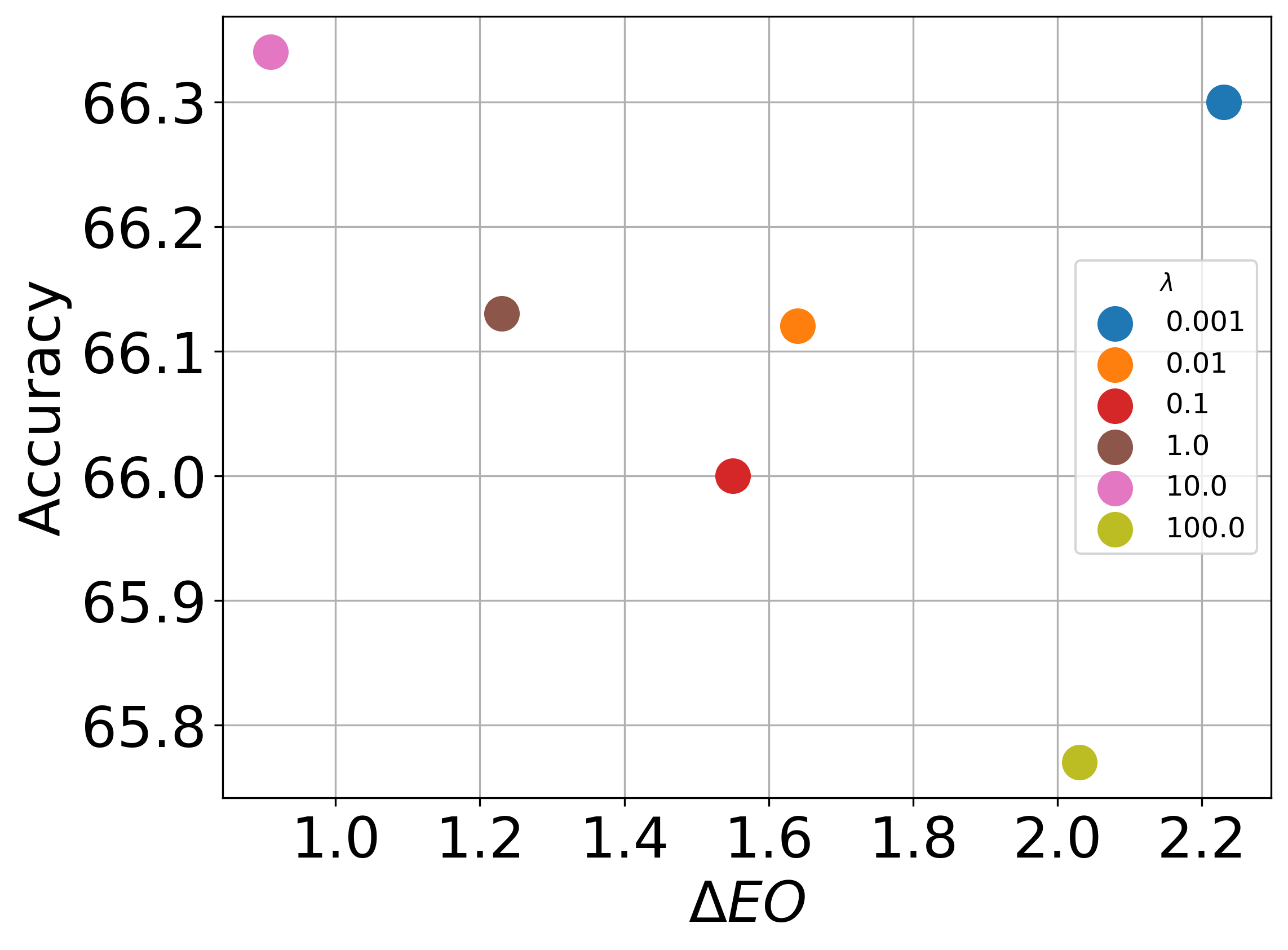}
\caption{($\lambda_X$, $\lambda_A$), Pokec-n}
\label{fig:Pokec_N_Acc_EO_lambda}
\end{subfigure}
\caption{Sensitivity analysis for our proposed FASD method in terms of Accuracy Vs. $\Delta \text{EO}$ on hyper-parameters $N_{\text{steps}}$ and ($\lambda_X$, $\lambda_A$): (a) $N_{\text{steps}}$, NBA; (b) $N_{\text{steps}}$, Pokec-z; (c) $N_{\text{steps}}$, Pokec-n; (d) ($\lambda_X$, $\lambda_A$), NBA; (e) ($\lambda_X$, $\lambda_A$), Pokec-z; (f) ($\lambda_X$, $\lambda_A$), Pokec-n.}

\label{fig:hyper_sen_2}
\vskip -0.15in
\end{figure}

\section{Hyper-Parameter Sensitivity}

To further investigate the impact of the hyper-parameters in our proposed FASD method, we present the results of additional sensitivity analyses on NBA, Pokec-z and Pokec-n datasets in Figure \ref{fig:hyper_sen_2}. Figures \ref{fig:NBA_Acc_EO_alpha}, \ref{fig:Pokec_Z_Acc_EO_alpha}, \ref{fig:Pokec_N_Acc_EO_alpha} depict the accuracy and equal opportunity of the proposed method as we vary the number of reverse-diffusion steps $N_{\text{steps}}$, while Figures \ref{fig:NBA_Acc_EO_lambda}, \ref{fig:Pokec_Z_Acc_EO_lambda}, \ref{fig:Pokec_N_Acc_EO_lambda} depict the accuracy and equal opportunity of the proposed method as we vary the hyper-parameters $\lambda_X$ and $\lambda_A$ which control the contribution of the fairness-based perturbations in the fairness-aware forward diffusion process. 

In the case of $N_{\text{steps}}$, the results clearly show, across all three datasets, that the ideal performance is achieved when the number of reverse-diffusion steps is relatively small ($\{2,3,4,5\}$). Large number of steps ($N_{\text{steps}} > 5$) leads to a drop in both model accuracy and fairness, while a single step ($N_{\text{steps}}=1$) yields high accuracy but with pronounced bias. 
This highlights that the proposed method requires more than one reverse-diffusion step to effectively debias the input subgraphs. 
Similarly, in the case $\lambda_X$ and $\lambda_A$ hyper-parameters, the results demonstrate that very small values ($\lambda_X$, $\lambda_A$ < $0.1$) yield limited improvement in equal opportunity, while very large values ($\lambda_X$, $\lambda_A$ > $10$) cause a significant drop in accuracy. Values within the range of $[0.1,10.0]$ lead to the optimal balance between accuracy and equal opportunity. In particular, $0.1$ leads to the best results in the case of NBA dataset, while $10.0$ lead to the best results on 
Pokec-z and Pokec-n dataset.

\end{document}